\documentclass[preprint,12pt,sort&compress]{elsarticle}

\usepackage[utf8]{inputenc} 
\usepackage[T1]{fontenc}    
\usepackage{hyperref}       
\usepackage{url}            
\usepackage{booktabs}       
\usepackage{amsfonts}       
\usepackage{nicefrac}       
\usepackage{microtype}      
\usepackage[flushleft]{threeparttable}
\usepackage{mathrsfs}

\usepackage{graphicx,color}
\usepackage{epsfig}
\usepackage{amsmath}
\usepackage{mathrsfs}
\usepackage{amssymb}
\usepackage{subfigure}
\usepackage{multirow}
\usepackage{enumitem}
\usepackage{footnote}
\usepackage{algpseudocode,algorithm}
\setlist[enumerate]{leftmargin=*}
\usepackage{setspace}
\usepackage{multirow}
\usepackage{verbatim}
\usepackage{amsthm}

\newtheorem{lem}{Lemma}
\newtheorem{prop}{Proposition}
\newtheorem{asum}{Assumption}
\newtheorem*{prof}{Proof}
\newtheorem*{profsketch}{Proof sketch}

\newtheorem{rem}{Remark}

\graphicspath{{figures/}}

\begin{document}
\setlength\tabcolsep{4.5pt} 

\begin{frontmatter}
\title{Root-cause Analysis for Time-series Anomalies via Spatiotemporal Graphical Modeling in Distributed Complex Systems}
\author{Chao Liu}
\author{Kin Gwn Lore}
\author{Zhanhong Jiang}
\author{Soumik Sarkar\corref{cor1}}
        \cortext[cor1]{Corresponding author}
        \ead{soumiks@iastate.edu}
\address{Department of Mechanical Engineering, Iowa State University, Ames, IA 50011, USA}

\begin{abstract}
  Performance monitoring, anomaly detection, and root-cause analysis in complex cyber-physical systems (CPSs) are often highly intractable due to widely diverse  operational modes, disparate data types, and complex fault propagation mechanisms. This paper presents a new data-driven framework for root-cause analysis, based on a spatiotemporal graphical modeling approach built on the concept of symbolic dynamics for discovering and representing causal interactions among sub-systems of complex CPSs. We formulate the root-cause analysis problem as a minimization problem via the proposed inference based metric and present two approximate approaches for root-cause analysis, namely the sequential state switching ($S^3$, based on free energy concept of a restricted Boltzmann machine, RBM) and artificial anomaly association ($A^3$, a classification framework using deep neural networks, DNN). Synthetic data from cases with failed pattern(s) and anomalous node(s) are simulated to validate the proposed approaches. Real dataset based on Tennessee Eastman process (TEP) is also used for comparison with other approaches. The results show that: (1) $S^3$ and $A^3$ approaches can obtain high accuracy in root-cause analysis under both pattern-based and node-based fault scenarios, in addition to successfully handling multiple nominal operating modes, (2) the proposed tool-chain is shown to be scalable while maintaining high accuracy, and (3) the proposed framework is robust and adaptive in different fault conditions and performs better in comparison with the state-of-the-art methods.
\end{abstract}

\begin{keyword}
Distributed complex system, anomaly detection, root cause analysis.
\end{keyword}
\end{frontmatter}

\section{Introduction}
With a large number of highly interactive sub-systems in cyber-physical systems (CPSs), performance monitoring, anomaly detection, and root-cause analysis is challenging in terms of widely diverse operational modes, disparate data types, and diverse fault mechanisms~\cite{bhuyan2016multi,sanislav2012cyber}. Data-driven approaches for anomaly detection in CPSs are more and more focused, especially in the situation that data (e.g., measurements, logs, events) are becoming increasingly accessible and abundant (in terms of cost and availability)~\cite{bhuyan2016multi,hu2014online,yuan2015online,jiang2011machine}. To exploit the large data availability in distributed CPSs and implement effective monitoring and decision making, data-driven approaches are being explored including (i) learning and inference in CPSs with multiple nominal modes, (ii) identifying anomaly and root cause without labeled data, and (iii) handling continuous and discrete data simultaneously~\cite{zhao2014adaptive,KST14}. Moreover, the data-driven methods for complex systems need to be robust, flexible, scalable and adaptive~\cite{xia2015novel,erfani2016high}.

For the anomaly detection with multivariate time-series, intensive research has been implemented with proposed approaches such as clustering \cite{aghabozorgi2015time,moshtaghi2016online}, fuzzy c-means \cite{izakian2014anomaly}, sparse representation \cite{takeishi2014anomaly}, Hidden Markov models \cite{li2017multivariate}, multivariate orthogonal space transformation \cite{serdio2014fault}, and independent component analysis \cite{palmieri2014distributed}. The temporal information is more analyzed in exploring the multivariate time-series, and the spatial relationship is recently applied for anomaly detection \cite{hu2007multivariate,qiu2012granger}.

In order to support decision making in CPSs~\cite{azevedo2016learning}, root-cause analysis is essential after an anomaly is detected. Here, root-cause analysis is intended to find out the anomalous variable(s) or sub-system(s) among all of the measured variables or sub-systems that cause the anomaly of the system (or the one or more time-series in the multivariate time-series that cause the anomaly). Among the root-cause analysis approaches, classification methods are widely used in fault localization~\cite{sen2016binarization}, while graphical modeling approaches (e.g., Bayesian network~\cite{xiang2014acquisition}) are becoming more and more applied in understanding of local structures to identify the root cause. One important aspect is that complex interactions between sub-systems are fundamental characteristics in distributed CPSs. These interactions can be used for modeling the system behavior and infer the root cause of an anomaly~\cite{sanislav2012cyber}. In this context, causality, as a tool to determine the cause and the effect in systems (i.e., analyze the interactions between the sub-systems) ~\cite{pearl2009causality}, is naturally suited for root-cause analysis in CPSs. Studies show that the causality from and to the fault variable presents differences and hence can be used to reason the root-cause from multivariate time series data~\cite{yuan2014root,mori2014identification,landman2014fault,chattopadhyay2014causality}.
\cite{qiu2012granger} derived neighborhood similarity and coefficient similarity from Granger-Lasso algorithm, to compute anomaly score and ascertain threshold for anomaly detection and infer the faulty variable. A causality analysis index based on dynamic time warping is proposed by Li et. al.~\cite{li2015dynamic} to determine the causal direction between pairs of faulty variables in order to overcome the shortcoming of Granger causality in nonstationary time series. Fault-related variables can be clustered and root-cause analysis is implemented in each cluster. A dynamic uncertain causality graph is proposed for probabilistic reasoning and applied to fault diagnosis of generator system of nuclear power plant~\cite{zhang2014dynamic}. The proposed approaches provide efficient tools in discovering causality in complex systems, while an approach for interpreting the variation of causality into decisions on failed patterns of fault variable/node/time-series is less investigated. Here, a pattern represents the individual behavior of or the causal dependency between the variables/nodes in the system and a failed pattern means that the behavior/dependency is significantly changed under an anomalous condition.

Data-driven approaches that can detect anomaly and infer root cause without using labeled training data (i.e., with data pre-classified as being anomalous, or as each pre-classified anomalous condition) are critical for applications in distributed complex CPSs~\cite{jiang2017energy,liu2017bridge,liu2018multivariate,yang2018unsupervised}, as collecting labeled data in these systems is time-consuming and some of the anomalies cannot be experimentally injected or easily collected from field operation. During the last decade, unsupervised approaches have received a large amount of attention, where the assumption regarding the ratio of the amount of anomalous data to nominal data is different for different problems and requires significant prior knowledge about the underlying system~\cite{song2013toward}. To implement unsupervised anomaly detection for distributed complex systems, an energy-based spatiotemporal graphical modeling framework~\cite{LGJS17cps}, where a spatiotemporal feature extraction scheme is built on Symbolic Dynamic Filtering (SDF)~\cite{gupta2008fault,RRSY09} for discovering and representing causal interactions among the sub-systems (defined as spatiotemporal pattern network, STPN) and the characteristics of the features in nominal condition are captured by restricted Boltzmann machine (RBM)~\cite{HS06}. With this framework (noted as STPN+RBM), the main intent is to capture various system-level nominal patterns such that anomaly detection can be carried out without using labeled data (specifically without the need of data examples for pre-classified anomalies).

Based on the formulation presented in \cite{cdc17data}, this work provides the proof of the concept with detailed description of the problem and presents a semi-supervised tool for root-cause analysis in complex systems with pattern-based root-cause analysis algorithm and a novel node-based inference algorithm to isolate the possible failed node (variable or subsystem) in the distributed complex systems. Two approaches for root-cause analysis, namely the \textit{sequential state switching} ($S^3$, based on free energy concept of an RBM) and \textit{artificial anomaly association} ($A^3$, a supervised framework using deep neural networks, DNN) are presented in \cite{cdc17data}. As $S^3$ and $A^3$ are built on the interactions (patterns in STPN) between sub-systems of CPSs, they are naturally pattern-based, thus we present a node-inference method in this work to interpret the faulty node (variable or sub-system in CPSs). Synthetic data from cases with failed pattern(s) and faulty node are simulated to validate the proposed approaches. Real data based on Tennessee Eastman process (TEP) is used for testing the proposed approach. The results are compared with the state-of-the-art methods in terms of root-cause analysis performance.

The contributions of this work include: (i) framing the root-cause analysis problem as a minimization problem using the inference based metric as well as the proof of the concept, (ii) two new approximate algorithms--$S^3$ and $A^3$--for the root-cause analysis problem with node inference algorithm to isolate the possible failed node (variable or subsystem), (iii) validation of the proposed algorithms in terms of accuracy, scalability, robustness, adaptiveness and efficiency using synthetic data sets and real data sets under a large number of different fault scenarios, and (iv) performance comparison with the state-of-the-art methods.

The remaining sections of the paper are organized as follows. Section \ref{secSDF} provides brief background and preliminaries including definition of STPN, basics of RBM, and STPN+RBM framework for anomaly detection; Section \ref{secMethod} presents the problem formulation of the root-cause analysis algorithm, the proposed approaches--$S^3$ and $A^3$--for root-cause analysis, and the node-inference approach to identify the faulty node; Section \ref{secResults} describes results for validation and comparisons of the proposed approaches with the state-of-the-art methods using synthetic and real datasets. Finally, the paper is summarized and concluded with directions for future work in Section \ref{secConclusion}.

\section{Background and preliminaries}
\label{secSDF} 
Graphical modeling is increasingly being applied in complex CPSs for learning and inference \cite{sarkar2016pgm}. Among different types of graphs, \emph{directed acyclic graph} (DAG) is typical and widely discussed, as the acyclicity assumption simplifies the theoretical analysis and is reasonable in usual cases \cite{eberhardt2007interventions}. However, the assumption is also not satisfied in many other cases, when causal cycles exist. An example is that causal cycles occur frequently in biological systems such as gene regulatory networks and protein interaction networks \cite{mooij2011causal}. Without the acyclicity assumption, causality with cycles (named as directed cyclic graph and directed pseudograph) is more difficult to be fully discovered \cite{lacerda2008discovering}. But the interpretation of cycles and loops may significantly improve the quality of the inferred causal structure \cite{mooij2011causal}. Several algorithms are proposed to discover causality in cyclic cases \cite{hyttinen2012learning,lacerda2008discovering,itani2010structure}, but fully discovering the graphical models with cycles is still challenging and the discovered graph may be inaccurate.

In this case, our previous work presented STPN+RBM framework for discovering the graphs that can be reasonably applied in modeling CPSs, where STPN is treated as a weaker learner to extract features (e.g., interactions and causality) between sub-systems, and RBM is used as a boosting approach to capture multiple operation modes in CPSs.

\subsection{STPN}
\label{secSTPN}
Data abstraction (partitioning/discretization) is the first step in STPN modeling, followed by learning Markov machines (defined by $D$-Markov machine and $xD$-Markov machine in SDF). While details can be found in~\cite{sarkar2014sensor,LGJS16conf}, a brief description is provided as follows for completeness.

Let $X=\{X^{\mathbb{A}}(t)$, $t \in \mathbb{N}$, $\mathbb{A}=1,2,\cdots, f\}$ be a multivariate time series, where $f$ is the number of variables or dimension of the time series. Let $\mathbb{X}$ denote a set of partitioning/discretization functions~\cite{RRSY09}, $\mathbb{X}: \ X(t)\to S$, that transforms a general dynamic system (time series $X(t)$) into a symbol sequence $S$ with an alphabet set $\Sigma$. Various partitioning approaches have been proposed in the literature, 
such as uniform partitioning (UP), maximum entropy partitioning (MEP, used for the present study), 
maximally bijective discretization (MBD),  
 and statistically similar discretization (SSD)~\cite{SS16}.
Subsequently, a probabilistic finite state automaton (PFSA) is employed to describe states (representing various parts of the data space) and probabilistic transitions among them (can be learned from data) via $D$-Markov and $xD$-Markov machines.

An STPN is defined as:

\textbf{Definition}. A PFSA based STPN is a 4-tuple $W_{D} \equiv ( Q^{a}, \Sigma^{b}, \Pi^{ab}, \Lambda^{ab})$: (a, b denote nodes of the STPN)
\begin{enumerate} 
  \item $Q^{a}=\{q_{1}, q_{2}, \cdots, q_{|Q^{a}|}\}$ is the state set corresponding to symbol sequences ${S^{a}}$.  
  \item $\Sigma^{b}=\{\sigma_{0}, \cdots, \sigma_{|\Sigma^{b}|-1}\}$ is the alphabet set of symbol sequence ${S^{b}}$. 
  \item $\Pi^{ab}$ is the symbol generation matrix of size $|Q^{a}| \times |\Sigma^{b}|$, the $ij^{th}$ element of $\Pi^{ab}$ denotes the probability of finding the symbol $\sigma_{j}$ in the symbol string  ${s^{b}}$ while making a transition from the state $q_{i}$ in the symbol sequence ${S^{a}}$; while self-symbol generation matrices are called atomic patterns (APs) i.e., when $a=b$, cross-symbol generation matrices are called relational patterns (RPs) i.e., when $a \neq b$.  
  \item $\Lambda^{ab}$ denotes a metric that can represent the importance of the learned pattern (or degree of causality) for $a \rightarrow b$ which is a function of $\Pi^{ab}$.
\end{enumerate}

\subsection{RBM}\label{sec:rbm}
As an energy-based method, weights and biases of RBM are learned to obtain low energy (or high probability) of the features observed from training data, which is the basis of an energy-based model~\cite{HS06}. If the training data is (mostly) nominal in operating conditions, the learned RBM should capture the nominal behavior of the system. 

Consider a system state that is described by a set of visible variables $\textbf{v} = (v_1, v_2, \cdots, v_D)$ and a set of hidden (latent) variables $\textbf{h} = (h_1, h_2, \cdots, h_F)$. With this configuration, the probability of a state $P(\textbf{v}, \textbf{h})$ is only determined by the energy of the configuration (\textbf{v}, \textbf{h}) and follows the Boltzmann distribution,
\begin{equation}
  P(\textbf{v}, \textbf{h}) = \frac{e^{-E(\textbf{v}, \textbf{h})}}{\sum_{\textbf{v}, \textbf{h}}e^{-E(\textbf{v}, \textbf{h})}}
\end{equation}
Typically, during training, weights and biases are obtained via maximizing likelihood of the training data.

Considering the weak learner with STPN in the interpretation of causality within distributed complex systems, RBM is applied as a boosting approach to form a strong learner based on multiple STPNs. Also, RBM is used for capturing multiple operating modes in complex CPSs.

\subsection{STPN+RBM framework}
\label{subsecstpnrbm}
The steps of learning the STPN+RBM model are \cite{liu2017unsupervised}:
\begin{enumerate}
\item Learning APs and RPs (atomic patterns, individual node behaviors; and relational patterns, pair-wise interaction behaviors) from the multivariate training symbol sequences. 
\item Considering short symbol sub-sequences from the training sequences and evaluating $\Lambda^{ab}\ \forall a,b$ for each short sub-sequence, noted as the inference based metric $\Lambda^{ab}$ for the pattern $a\to b$. 
\item For one sub-sequence, based on a user-defined threshold on $\Lambda^{ij}$, assigning state $0$ or $1$ for each AP and RP; thus every sub-sequence leads to a binary vector of length $L$, where $L = \text{number of APs}~(\#AP) + \text{number of RPs}~(\#RP)$. 
\item An RBM is used for capturing system-wide behavior with nodes in the visible layer corresponding to learnt APs and RPs. 
\item The RBM is trained with binary vectors generated from nominal training sub-sequences. 
\item Anomaly detection is carried out by estimating the probability of occurrence of a test STPN pattern vector via trained RBM.
\end{enumerate}

Root-cause analysis is based on this framework in two aspects: (i) the inference based metric is used for formulating root-cause analysis algorithm, and (ii) anomaly detection status of the complex CPS with STPN+RBM is the trigger to implement the root-cause analysis.

\section{Methodology}
\label{secMethod}
\subsection{Root-cause analysis problem formulation}
\label{subsecformulation}
With the definition of STPN in Section \ref{secSTPN} and STPN+RBM framework in Section \ref{subsecstpnrbm}, the inference based metric is employed for evaluating the patterns (APs \& RPs). The inference based metric computation includes a modeling phase and a inference phase \cite{LGJS16conf}. In the modeling phase, the time-series in the nominal condition is applied, noted as $X=\{X^{\mathbb{A}}(t)$, $t \in \mathbb{N}$, $\mathbb{A}=1,2,\cdots, f \}$, where $f$ is the number of variables or the dimension of the time series. The time series is then symbolized into $S=\{S^{\mathbb{A}}\}$  using MEP followed by the generation of state sequence using the STPN formulation where $Q=\{Q^a,\ a=1,2,\cdots, f \}$. In the learning phase, short time-series is considered, $\tilde{X}=\{\tilde{X}^{\mathbb{A}}(t)$ for the short time-series in nominal condition, $t \in \mathbb{N}^{*}$, $\mathbb{A}=1,2,\cdots, f \}$, where $\mathbb{N}^{*}$ is a subset of $\mathbb{N}$, and $\hat{X}=\{\hat{X}^{\mathbb{A}}(t)$ for the short time-series in anomalous condition. The corresponding short symbolic subsequences is noted as $\tilde{S}=\{\tilde{S}^{\mathbb{A}}\}$ and $\hat{S}=\{\hat{S}^{\mathbb{A}}\}$ for nominal condition and anomalous condition respectively, and the state sequences $\tilde{Q}$ and $\hat{Q}$ correspondingly.

We define an importance metric $\Lambda^{ab}$ for a given short subsequence (described by short state subsequence $\tilde{Q}$ and short symbol subsequence $\tilde{S}$). The value of this metric suggests the importance of the pattern $a \rightarrow b$ as evidenced by the short subsequence. In this context, we consider
\begin{equation}
\Lambda^{ab}(\tilde{Q}, \tilde{S}) \propto Pr(\{\tilde{Q}^{a},\tilde{S}^b\}|\{Q^{a},S^b\})
\end{equation}
where $Pr(\{\tilde{Q}^{a},\tilde{S}^b\}|\{Q^{a},S^b\})$ is the conditional probability of the joint state-symbol subsequence given the sequence from the modeling phase.

With this definition of $\Lambda^{ab}$ and with proper normalization, the inference based metric $\Lambda^{ab}(\tilde{Q}, \tilde{S})$ can be expressed as follows according to \cite{LGJS16conf},
\begin{align}
\begin{split}
& \Lambda^{ab}(\tilde{Q}, \tilde{S})  =\\
  &   K \prod_{m=1}^{|Q^{a}|}{\frac{(\tilde{N}^{ab}_{m})!(N^{ab}_{m}+|\Sigma^{b}|-1)!}{(\tilde{N}^{ab}_{m}+N^{ab}_{m}+|\Sigma^{b}|-1)!}}
                      \prod_{n=1}^{|\Sigma^{b}|}{\frac{(\tilde{N}^{ab}_{mn}+N^{ab}_{mn})!}{(\tilde{N}^{ab}_{mn})!(N^{ab}_{mn})!}}
\label{equProbOnline1}
\end{split}
\end{align}
where, $K$ is a proportional constant, $N_{mn}^{ab} \triangleq |\{(Q^{a}(k),S^{b}(k+1): S^{b}(k+1) = \sigma^{b}_n\ | \ Q^{a}(k) = q^a_m \}|$ is to get the number of the state-symbol pairs in the set $\{(Q^{a}(k),S^{b}(k+1)\}$ where $S^{b}(k+1) = \sigma^{b}_n$ given $Q^{a}(k) = q^a_m$, $N_{m}^{ab}=\sum_{n=1}^{|\Sigma^{b}|}(N_{mn}^{ab})$, $\tilde{N}^{ab}_{mn}$ and $\tilde{N}^{ab}_{m}$ are similar to $N_{mn}^{ab}$ and $N_{m}^{ab}$, $|Q^{a}|$ is number of states in state sequence $\tilde{Q}$, and $|\Sigma^{b}|$ is number of symbols in symbol sequence $\tilde{S}$.

A detailed derivation can be found in \cite{LGJS16conf}.

For an anomalous condition, the inference based metric $\Lambda^{ab}(\hat{Q}, \hat{S})$ for a pattern $a\to b$ is obtained as follows\cite{LGJS16conf},

\begin{align}
\begin{split}
& \Lambda^{ab}(\hat{Q}, \hat{S})  =\\
  &   K \prod_{m=1}^{|Q^{a}|}{\frac{(\hat{N}^{ab}_{m})!(N^{ab}_{m}+|\Sigma^{b}|-1)!}{(\hat{N}^{ab}_{m}+N^{ab}_{m}+|\Sigma^{b}|-1)!}}
                      \prod_{n=1}^{|\Sigma^{b}|}{\frac{(\hat{N}^{ab}_{mn}+N^{ab}_{mn})!}{(\hat{N}^{ab}_{mn})!(N^{ab}_{mn})!}}
\label{equProbOnline1}
\end{split}
\end{align}
where $\hat{N}$ is with the similar definition in Eq. \ref{equProbOnline1}, while it is emanated from the time series $\hat{X}=\{\hat{X}^{\mathbb{A}}(t)$, $t \in \mathbb{N}^{*}$, $\mathbb{A}=1,2,\cdots, f \}$ in the anomalous condition.

For simplicity, we denote the metric in the nominal condition and the anomalous condition as  $\Lambda^{ab}_{nom}$ and $\Lambda^{ab}_{ano}$ respectively. Let $\delta \Big(\ln( \Lambda ^{ab}) \Big)$ denote the variation of the metric due to the presence of an anomaly,
\begin{align}
\begin{split}
\delta \Big( \ln( \Lambda ^{ab}) \Big)= \ln\big(\Lambda^{ab}_{nom}\big) - \ln\big(\Lambda^{ab}_{ano}\big).
\label{equdeltadef}
\end{split}
\end{align}

We also define the set of metric values for all patterns in the nominal condition as $\Lambda_{nom} = \{\Lambda^{ab}_{nom}\} \ \forall a, b$ and the set of metric values for all patterns in the anomalous condition as $\Lambda_{ano} = \{\Lambda_{ano}^{ab}\} \ \forall a, b$.

To explain the changes of metric $\Lambda$ caused by the anomalous condition, we assume a two-state case (i.e., the $Q$ only contains two states $q_{1}$ and $q_2$), the short state sequences between the nominal condition and anomalous condition (with two subsystems $a$ and $b$) are $\tilde{Q}=\{q_{1}^{a} q_{1}^{a} \cdots q_{2}^{a};\ q_{1}^{b} q_{1}^{b} \cdots q_{2}^{b}\}$ and $\hat{Q}=\{q_{1}^{a} q_{2}^{a} \cdots q_{2}^{a};\ q_{1}^{b} q_{1}^{b} \cdots q_{2}^{b}\}$ respectively.

\begin{figure}[htbp]
  \centering
  \includegraphics[trim={0 0 0 0},scale=0.36]{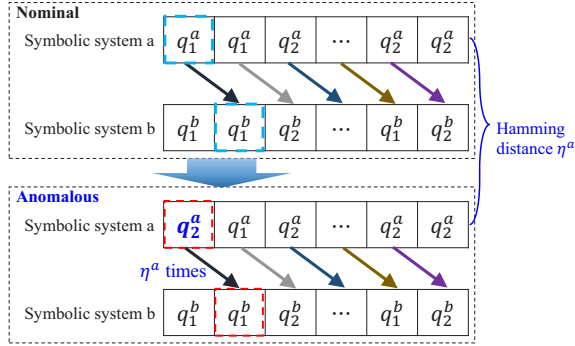}
  \caption{A two-state case of anomalous condition with two sub-systems. The state transitions between the subsystems $a$ and $b$ are first defined in the nominal condition (shown in the top panel) assuming the depth $D=1$ and the time lag $p=1$. Then, in the anomalous condition, changes occur from the state $q_1^a$ to the state $q_2^a$ in the subsystem $a$ and we assume that the changes only exist in the transitions $q_1^a\to q_1^b$ (i.e., they change to $q_2^a\to q_1^b$ due to the anomaly). The Hamming distance between the sequence for the subsystem $a$ in the anomalous condition and that in the nominal condition is $\eta^a$.}
  \label{figanom_demo}%
\end{figure}

An illustration of the changes in the anomaly is shown in Fig. \ref{figanom_demo}. Note that for simplicity, we assume the depth $D=1$. Hence, the state $q$ and symbol $\sigma$ are equivalent, and the state generation matrix defined in Section \ref{secSTPN}, can be used for state-to-state transition computation.

\begin{asum}\label{asum1}
In this illustrative example, suppose that the states change from $q_{1}^{a}$ in the nominal condition to $q_{2}^{a}$ in the anomalous condition in subsystem $a$ for $\eta^a$ times, and the changes only occur in the state transitions from $q_1^a\to q_1^b$ for the x$D$-Markov machine.
\end{asum}

From Assumption \ref{asum1}, the Hamming distance between $\tilde{Q}$ and $\hat{Q}$ is $\eta^a \in \mathbb{N}$. Note, in a real case, this may be caused by the magnitude increase of a real valued observation in the anomalous condition. However, the practical situation is typically more complicated, and the changes may occur in different state transitions at multiple time periods simultaneously.

Based on the assumption, we have,
\begin{equation}
\hat{N}^{ab}_{m} = \tilde{N}^{ab}_{m}-sgn(q_m)\eta^{a}
\end{equation}
\begin{equation}
sgn(q_x) =
    \begin{cases}
      1 &\text{if}\quad q_x=q_1\\
     -1 &\text{if}\quad q_x=q_2
    \end{cases}
\end{equation}
where the states $q_1$ and $q_2$ are from the state set $Q$ in the state sequence. Note that $\tilde{N}_{m}$ depend on the partitioning process.

From Assumption \ref{asum1}, the changes only exist from $q_1^a\to q_1^b$ to $q_2^a\to q_1^b$ (i.e., $q_1^a\to \sigma_1^b$ to $q_2^a\to \sigma_1^b$ with $D=1$). Therefore, $\hat{N}_{m1}^{ab}$ is affected, where $\tilde{N}_{m1}^{ab} \triangleq |\{(\hat{Q}^{a}(k),\hat{S}^{b}(k+1): \hat{S}^{b}(k+1) = \sigma^{b}_1\ | \ \hat{Q}^{a}(k) = q^a_m \}|$. $\hat{N}_{m2}^{ab}$ doesn't get affected. Thus,
\begin{equation}
\begin{aligned}
&\hat{N}^{ab}_{m1} = \tilde{N}^{ab}_{m1}-sgn(q_m)\eta^{a}, \quad
&\hat{N}^{ab}_{m2} = \tilde{N}^{ab}_{m2}.
\end{aligned}
\end{equation}

\begin{asum}\label{asum2}
In this analysis, for simplicity, we assume $N_m^{ab}=k\tilde{N}_m^{ab}, m=1,2$, where the short sub-sequence is a subset of the sequence from the modeling phase, and with the same features.  $k\gg1$, when the sequence in the modeling phase (historical data in the nominal condition) is long enough.

\end{asum}
Similarly, the consequence of Assumption \ref{asum2} can get $N_{mn}^{ab}=k\tilde{N}_{mn}^{ab},\ m,n=1,2$.

\begin{asum}\label{asum3}
With proper partitioning, $\tilde{N}_{11}$ and $\tilde{N}_{21}$ are within similar orders.
\end{asum}
\begin{lem}\label{lem1}
Let Assumption \ref{asum3} hold and $\tilde{N}_1^{ab}=t_1\tilde{N}_{11}^{ab}$ such that $(t_1-1)k>1$.
\end{lem}
\begin{profsketch}
We have Assumption 3 such that there exists a constant $\zeta>0$ to enable $t-1>\zeta$, which combined with Assumption 2 yields the desired result.
See Section \ref{proof1} in the supplementary material.
\end{profsketch}

Similarly, let $\tilde{N}_2^{ab}=t_2\tilde{N}_{21}^{ab}$. With this setup, we have the following proposition,
\begin{prop}
Let Assumptions \ref{asum1}-\ref{asum3} hold. The variation of the metrics in the anomalous condition $\delta \Big(\ln( \Lambda ^{ab}) \Big)>0$, when $\eta^a\geq1$.
\end{prop}
\begin{profsketch}
\nonumber
To show the conclusion, we first give the explicit expression of $\delta \Big( \ln( \Lambda ^{ab}) \Big)$ using Eq. 5 in terms of the summation of logarithmic Gamma functions. According to the derivative property of the logarithmic Gamma function, the derivative of $\delta \Big( \ln( \Lambda ^{ab}) \Big)$ with respect to $\eta^a$ can be represented by a summation of the Harmonic series, which can also be approximated by a logarithmic function plus a Euler-Mascheroni constant. Eventually, the derivative of $\delta \Big( \ln( \Lambda ^{ab}) \Big)$ with respect to $\eta^a$ is expressed as a summation of several logarithmic functions, which are evaluated positive by combining Lemma 1 and definitions aforementioned to get the conclusion.
See Section \ref{proof2} in the supplementary material.
\end{profsketch}

Therefore, the variation of the metrics $\delta \big(\ln(\Lambda ^{ab})\big)$ is relative to the changes of states $\eta^a$ in the anomalous condition. Based on this metric, the root-cause analysis algorithm can be formulated.

The main idea behind the presented root-cause analysis algorithm is to perturb the distribution of testing patterns in an artificial way to make it close to the distribution of nominal patterns, to identify the possible nodes/patterns which bring the test pattern distribution sufficiently close to the distribution of nominal patterns. Then the identified nodes/patterns can be determined as the root-cause(s) for the detected anomaly. More formally, let us assume that an inference metric $\Lambda_{ano}^{ab}$ changes to $\Lambda_{ano}^{ab\prime}$ due to an artificial perturbation in the pattern $a\to b$. We now consider a subset of patterns, for which we have the set of inference metrics $\{\Lambda_{ano}^{ab}\} \subset \Lambda_{ano}$. Let a perturbation in this subset changes the overall set of metrics to $\Lambda_{ano}^{\prime}$. Therefore, we have the following:
\begin{equation}
\Lambda_{ano}^{\prime}=
    \begin{cases}
      \Lambda_{ano}^{ab} \ \ \text{if}\ \  \Lambda_{ano}^{ab} \not\in \{\Lambda_{ano}^{ab}\}\\
      \Lambda_{ano}^{ab\prime} \ \ \text{if}\ \  \Lambda_{ano}^{ab} \in \{\Lambda_{ano}^{ab}\}
    \end{cases}
\end{equation}

The root cause analysis is then framed as a minimization problem between the set of nominal inference metrics $\Lambda$ and the set of perturbed inference metrics $\Lambda^{\prime}$ that can be expressed as follows:
\begin{align}
\begin{split}
\{\Lambda_{ano}^{ab}\}^{\star} = \min_{\{\Lambda_{ano}^{ab}\}} \mathscr{D}\Big(\Lambda_{ano}^{\prime},\Lambda_{nom}\Big),
\label{equInference}
\end{split}
\end{align}
where $\mathscr{D}$ is a distance metric (e.g., Kullback-Leibler Distance--KLD\cite{kullback1951information,frey2005comparison}) to estimate the distance between $\Lambda_{nom}$ and $\Lambda_{ano}^{\prime}$. The nodes (e.g., $a$ or $b$) or patterns (e.g., $a\to b$) involved in the optimal subset $\{\Lambda_{ano}^{ab}\}^{\star}$ will be identified as possible root cause(s) for the detected anomaly. However, exact solution for this optimization problem is not computationally tractable for large systems. Therefore, we propose two approximate algorithms: the sequential state switching ($S^3$) - a sequential suboptimal search method and artificial anomaly association ($A^3$) - a semi-supervised learning based method.

\subsection{Sequential state switching ($S^3$)}
The basic idea of $A^3$ is illustrated in \cite{cdc17data}, one-layer RBM is applied in this work to capture the system-wide behavior of the CPS, where the number of hidden units is chosen by maximizing the activation of the inputs from 16 to 256.

With the weights and biases of RBM, free energy can be computed. Free energy is defined as the energy that a single visible layer pattern would need to have in order to have the same probability as all of the configurations that contain $\textbf{v}$~\cite{hinton2012practical}, which has the following expression:
\begin{equation}
F(v)=-\sum_{i}{v_{i}a_{i}}-\sum_{j}\ln(1+e^{b_{j}+\sum_{i}{v_{i}w_{ij}}})
\label{equFreEngy}
\end{equation}
The free energy in nominal conditions is noted as $\tilde{F}$. In cases where there are multiple input vectors with more than one nominal mode, free energy in the nominal states can be averaged or used in conjunction with other metrics. In anomalous conditions, a failed pattern will shift the energy from a lower state to a higher state. Assume that the patterns can be categorized into two sets, $\textbf{v}^{nom}$ and $\textbf{v}^{ano}$.
By flipping the set of anomalous patterns $\textbf{v}^{ano}$, a new expression for free energy is obtained:
\begin{align}
\begin{split}
F^{s}(v)=&-\sum_{g}{v_{g}a_{g}}-\sum_{j}\ln(1+e^{b_{j}+\sum_{g}{v_{g}w_{gj}}})\\
           &-\sum_{h}{v^{\star}_{h}a_{h}}-\sum_{j}\ln(1+e^{b_{j}+\sum_{h}{v^{\star}_{h}w_{hj}}}), \\ &\quad \{v_{g}\} \in v^{nom}, \{v^{\star}_{h}\} \in  v^{\star,ano}
\label{equFreEngyAno}
\end{split}
\end{align}

Here, $v^{\star}$ has the opposite state to $v$ and represents that the probability of the pattern has been significantly changed. In this work, the probabilities of the patterns are binary (i.e. 0 or 1). Hence, we have that $v^{\star}=1-v$. The sequential state switching is formulated by finding a set of patterns $v^{ano}$ via $\min(F^{s}(v^{ano},v^{nom})-\tilde{F})$.

The root-cause analysis procedure is triggered when an anomaly is detected by STPN+RBM framework. With Algorithm \ref{algthmS3}, the threshold of free energy of RBM $F^c$ is initialized as the one in the anomalous condition, the candidate set of anomalous inputs $\textbf{v}_p$ are obtained with $\{v: F^{s}(\textbf{v})<F^{c}\}$, then at each step: (i) $\textbf{v}^{ano}$ is updated by adding the input $v_i$ with maximum free energy decrease during perturbations on the set $\textbf{v}^{nom}$, where $\textbf{v}^{ano}$ is the set of inputs with anomalous patterns (initially empty set); (ii) the set of inputs with nominal patterns $\textbf{v}^{nom}$ is updated by removing the input $v_i$ which is determined as anomalous at substep (i); (iii) the threshold of free energy of RBM $F^c$ is updated by the inputs $\textbf{v}^{\star, ano} \cup \textbf{v}^{nom}$, where $\textbf{v}^{\star, ano}$ is the inputs after perturbations; and (iv) the candidate set of inputs $\textbf{v}_p$ (i.e., the inputs of RBM) is updated by removing the input $v_i$.

An algorithmic description of this process (Algorithm \ref{algthmNode}) is presented as follows.

\begin{algorithm*}[!h]
\caption{Root-cause analysis with sequential state switching ($S^3$) method}
\label{algthmS3}
\begin{algorithmic}[1]
    \Procedure{STPN+RBM modeling} {}\Comment{Algorithm 1 in~\cite{LGJS17cps}}
        \State Online process of computing likelihoods/probabilities of APs \& RPs
        \State Training RBM to achieve low energy state, using binary vectors from nominal subsequences
    \EndProcedure
    \Procedure{Anomaly detection} {}\Comment{Algorithm 2 in~\cite{LGJS17cps}}
        \State  Online anomaly detection via the probability of the current state via trained RBM
    \EndProcedure
    \Procedure {Root-cause analysis}{}
        \If{$Anomaly=True$}
            \State $F^{c}_{0} \leftarrow F^{s}(\textbf{v})$ \Comment $F^{c}$ is the current free energy with input vector $\textbf{v}=\textbf{v}^{nom} \cup \textbf{v}^{ano}$.
            \State $\mathbf{v}_{p} \leftarrow \{v: F^{s}(\textbf{v})<F^{c}\}$
            \State $\mathbf{v}^{ano}=\emptyset$, $\mathbf{v}^{nom}=\mathbf{v}$
            \While{$\mathbf{v}_{p} \neq \emptyset \lor \{v: F^{s}(\mathbf{v}^{\star,ano},\mathbf{v}^{nom})<F^{c}\} = \emptyset$}
                \State $\mathbf{v}^{ano} \leftarrow \mathbf{v}^{ano} \cup \{v_{i}: F^{s}(\mathbf{v}^{\star,ano}\cup v_{i}^{\star},\textbf{v}^{nom})=F^{c}\}$
                \State $\mathbf{v}^{nom} \leftarrow \mathbf{v}^{nom} \setminus \mathbf{v}^{ano} $
                \State $F^{c} \leftarrow \min(F^{s}(\textbf{v}^{\star,ano}\cup v_{i}^{\star},\textbf{v}^{nom}))$, $v_{i} \in \mathbf{v}_{p}$, $v_{i}^{\star}=1-v_{i}$, $\textbf{v}^{\star,ano}=1-\textbf{v}^{ano}$
                \State $\mathbf{v}_{p} \leftarrow \mathbf{v}_{p}\setminus \{v_{i}: F^{s}(\mathbf{v}^{\star,ano}\cup v_{i}^{\star},\textbf{v}^{nom})=F^{c}\}$
            \EndWhile
        \EndIf
        \State A bijective function: $f: \mathbf{\Lambda} \rightarrow \mathbf{v} $
        \State $\mathbf{\Lambda}^{ano} =f^{-1}(\mathbf{v}^{ano})$
        \State \textbf{return} $\mathbf{\Lambda}^{ano}$
    \EndProcedure
\end{algorithmic}
\end{algorithm*}

It should be noted that free energy $F$ is used in Algorithm 1 to be applied as the distance metric of the Eq. \ref{equInference}, and it can be used along with other metrics such as KLD. Using KLD alongside with free energy is particularly useful when the distribution of free energy is obtained with multiple sub-sequences. KLD may be more robust as it takes multiple sub-sequences into account because a persistent anomaly across the subsequences will cause a significant impact on KLD.

\subsection{Artificial anomaly association ($A^3$)}
Artificial anomaly association is based on a method proposed in~\cite{lore2015hierarchical} to solve a multi-label classification problem using convolutional neural networks (CNNs). Instead of inferring a single class from the trained model, the framework solves $n_{out}$ classification sub-problems if an output vector of length $n_{out}$ is required using the previously learned model. The implementation of this formulation requires only a slight modification in the loss function: for an output sequence with length $n_{out}$, the loss function to be minimized for a data set $\mathcal{D}$ is the negative log-likelihood defined as:
\begin{equation}
\begin{split}
& \ell_{total}(\theta=\{W,b\},\mathcal{D}) = \\
& - \sum_{j=1}^{n_p} \sum_{i=0}^{\mathcal{|D|}} \left[ \log{\left(P(Y=y^{(i)}|x^{(i)},W,b)\right)} \right]_j
\end{split}
\end{equation}

where $\mathcal{D}$ denotes the training set, $\theta$ is the model parameters with $W$ as the weights and $b$ for the biases. $y$ is predicted target vector whereas $x$ is the provided input pattern. The total loss is computed by summing the individual losses for each sub-label.

The input is prepared in an $n^2$-element vector with values of either 0 or 1 which is learnt by the STPN model. We desire to map the input vector to an output vector of the same length (termed as the \textit{indicator label}), where the value of each element within the output vector indicates whether a specific pattern is anomalous or not. For nominal modes, the input vector may be comprised of different combinations of 0's and 1's, and the indicator labels will be a vector of all 1's (where the value 1 denotes nominal). However, if a particular element $i$ within the input vector gets flipped, then the indicator label corresponding to the $i$-th position in the output vector will be flipped and switches from 1 (nominal) to 0 (anomalous). In this way, we can identify that the $i$-th pattern is anomalous. With this setup, a classification sub-problem (i.e. is this pattern norminal, or anomalous?) can be solved for each element in the output vector given the input data. One might argue that multi-label classification is unnecessary and adds to higher computational overhead. Although a single class label denoting which pattern has failed may work for a single anomalous pattern case, it is not sufficient for simultaneous multiple pattern failures.

\textbf{Neural network parameters:} Training data is generated from 6 modes including both nominal and artificial anomalous data. The number of available training data is dependent on the size of the system under study. The training data is further split into an equally-sized training set and validation set. Testing was performed on datasets generated from entirely different modes. The DNN comprises 1 to 3 layers of 1024 to 4096 hidden units each and trained with dropout fractions of 0.4 to 0.8 using ReLU activations. A batch size ranging from 50 to 4096 is used. All hyperparameters as described are chosen by cross-validation, and the set of optimal hyperparameters vary depending on the training data used. The training procedure employs the early-stopping algorithm where training stops when validation error ceases to decrease.

\subsection{Node inference via $S^{3}$ and $A^{3}$}
\label{subsecNodeinf}
The proposed approaches, $S^{3}$ and $A^{3}$, intend to implement root-causes analysis by finding out the failed patterns (namely pattern-based root-cause analysis). This kind of pattern failures occurs in cyber-physical systems, where cyber-attacks may only compromise interactions among sub-systems without directly affecting sub-systems. In the meantime, there are a lot of cases where failures may emerge in sub-systems. In such cases, the failures mostly break connections from and to the failed sub-system (treated as node in the probabilistic graphical models).  \color{black}With patterns found by $S^{3}$ or $A^{3}$, node inference (i.e., to find out the failed node in the graphical model and the anomalous variable/sub-system in the CPS) is presented here to the later case with faulty variable (sub-system) in CPSs (namely node-based root-cause analysis).

Based on the anomalous patterns $\{\Lambda^{ano}\}$ obtained by $S^{3}$ or $A^{3}$, node inference is to find a subset $\mathbf{\hat{X}}$ of $X=\{X^{\mathbb{A}}(t)$, $t \in \mathbb{N}$, $\mathbb{A}=1,2,\cdots, n\}$ that can interpret all of the failed patterns $\mathbf{\Lambda}^{ano}$. For example, a pattern $N_{i}\to N_{j}$ is identified as failed, and it indicates that the two nodes $(N_{i}, N_{j})$ are potentially failed. If multiple patterns from or to a node are detected as anomalous, the node is more reliable to be classified as anomalous. Thus, the node inference can be carried out via computing the anomaly score of each node. Also, as the failed patterns contribute differently to the system, weights of failed patterns are defined. For $S^{3}$, the weights ($\Delta=\{\delta\}$) of failed patterns can be formed by the difference of free energy with and without the failed pattern, also they can be computed by the KLD metric. For $A^{3}$, the weights can be obtained by counting the number of occurrence of the patterns in multiple testing outputs, or evaluate the probability of the activation of the outputs.

As the RBM inputs ($\mathbf{v}$) are generated by the patterns of STPN ($\mathbf{\Lambda}$, $\Lambda^{a,a}$ for AP and $\Lambda^{a,b}$ for RP), each RBM inputs can be attributed to a pattern in STPN. And each weight ($\sigma$) of the failed pattern ($\Lambda^{a,b}$) in STPN can be connected to two nodes (variables or sub-systems) in the CPSs, i.e., $\sigma(v)\rightarrow \sigma(\Lambda)$, $\sigma(\Lambda) \rightarrow \sigma(X^{a},X^{b})$. With this setup, the anomaly score for each node $X^{\mathbb{A}}$ is obtained by sum of the weights,
\begin{equation}
S(X^{\mathbb{A}})=\sum\sigma(X^{\mathbb{A}},X^{a})+\sum\sigma(X^{a},X^{\mathbb{A}}), \quad X^{a} \in \mathbf{X}.
\end{equation}

A sequential searching for node inference is applied by finding out the node with maximum anomaly score each time. At each step: (i) the anomalous patterns $\Lambda^{ano}$ are obtained by the bijection function $f^{-1}(v^{ano})$, (ii) the weights of the anomalous patterns  $\Lambda^{ano}$ are obtained via the set of inputs $v^{ano}$, (iii) the anomaly scores $S$ (node based) are computed with the weights of the anomalous patterns $\Lambda^{ano}$, where each pattern is connected to a node set $(X^a, X^b)$ (head and tail nodes), (iv) the set of anomalous nodes $\tilde{X}$ is updated via finding out the node $X^{\mathbb{A}}$ with maximum anomaly score, and (v) the anomalous patterns are updated by removing the ones that connected to the anomalous node $X^{\mathbb{A}}$ determined at substep (iv). An algorithmic description of this process (Algorithm \ref{algthmNode}) is presented as follows.

\begin{algorithm*}[!h]
\caption{Node inference based on $S^{3}$ and $A^{3}$}
\label{algthmNode}
\begin{algorithmic}[1]
    \Procedure{Root-cause analysis}{}
    \State Find out the set of failed patterns $\mathbf{\Lambda}^{ano}$ using $S^{3}$ or $A^{3}$.
    \EndProcedure
    \Procedure {Node inference}{}
        \State A bijective function: $f: \mathbf{\Lambda} \rightarrow \mathbf{v} $
        \State $\mathbf{\hat{X}} = \emptyset $, $\mathbf{\bar{\Lambda}}^{ano} \leftarrow \mathbf{\Lambda}^{ano}$
        \While {$\mathbf{\bar{\Lambda}}^{ano} \neq \emptyset$}
            \State $\mathbf{\bar{v}}^{ano} \leftarrow f(\mathbf{\bar{\Lambda}}^{ano})$
            \State $\Delta =\emptyset$
            \For {$v^{ano} \in\mathbf{\bar{v}}^{ano}$}
                \State $\Lambda^{ano} =f^{-1}(v^{ano})$
                \State $(X^{a},X^{b})\ obtained\ by\ \Lambda^{ano}$
                \State $v^{\star,ano} = 1-v^{ano}$
                \State $\Delta \leftarrow \Delta \cup \{\delta(X^{a},X^{b}): (F^{s}(v^{\star,ano},\mathbf{v}\setminus \{v^{ano}\})-F^{c}_{0})/F^{c}_{0} \}$ \Comment Only for $S^{3}$, $F^{c}_{0}$ obtained in Algorithm 1
                \State $\Delta \leftarrow \Delta \cup \{\delta(X^{a},X^{b}): p(v^{ano}) \}$ \Comment Only for $A^{3}$, $p(v^{ano})$ obtained from DNN outputs
            \EndFor
            \State $\mathbf{S}\leftarrow \{S(X^{\mathbb{A}}): \sum \delta(X^{\mathbb{A}},X^{a})+\sum \delta(X^{a},X^{\mathbb{A}}), \delta \in \Delta, X^{a} \in \mathbf{X} \}$
            \State $\mathbf{\hat{X}}  \leftarrow \mathbf{\hat{X}}  \cup \{X^{\mathbb{A}}: S(X^{\mathbb{A}}) =\max(\mathbf{S}) \}$  
            \State $\Psi =\emptyset$
            \For {$\Lambda^{ano} \in \mathbf{\bar{\Lambda}}^{ano}$}
                \State $\{X^{a},X^{b}\} \ obtained\ by\ \Lambda^{ano}$
                \If{$\exists \hat{X}^{\mathbb{A}} \in \{X^{a},X^{b}\} , \hat{X}^{\mathbb{A}} \in \mathbf{\hat{X}}$}
                    $\Psi \leftarrow \Psi \cup \Lambda^{ano}$
                \EndIf
            \EndFor
            \State $\mathbf{\bar{\Lambda}}^{ano} \leftarrow \mathbf{\bar{\Lambda}}^{ano} \setminus \Psi$
        \EndWhile
    \State \textbf{return} $\hat{\mathbf{X}}$
    \EndProcedure
\end{algorithmic}
\end{algorithm*} 

Note that Algorithm \ref{algthmNode} intends to find out the minimum subset of nodes that can interpret all of the failed patterns obtained by Algorithm \ref{algthmS3}. Further analysis is being carried out including introducing penalty factor to optimize node searching that can possibly cause the patterns presented as failed and identify the anomaly degree of a specific node.

\section{Results and discussions}
\label{secResults}

\begin{figure*}[htbp]
  \centering
  \subfigure[Mode 1]{\includegraphics[trim={0 0 0 20},scale=0.29]{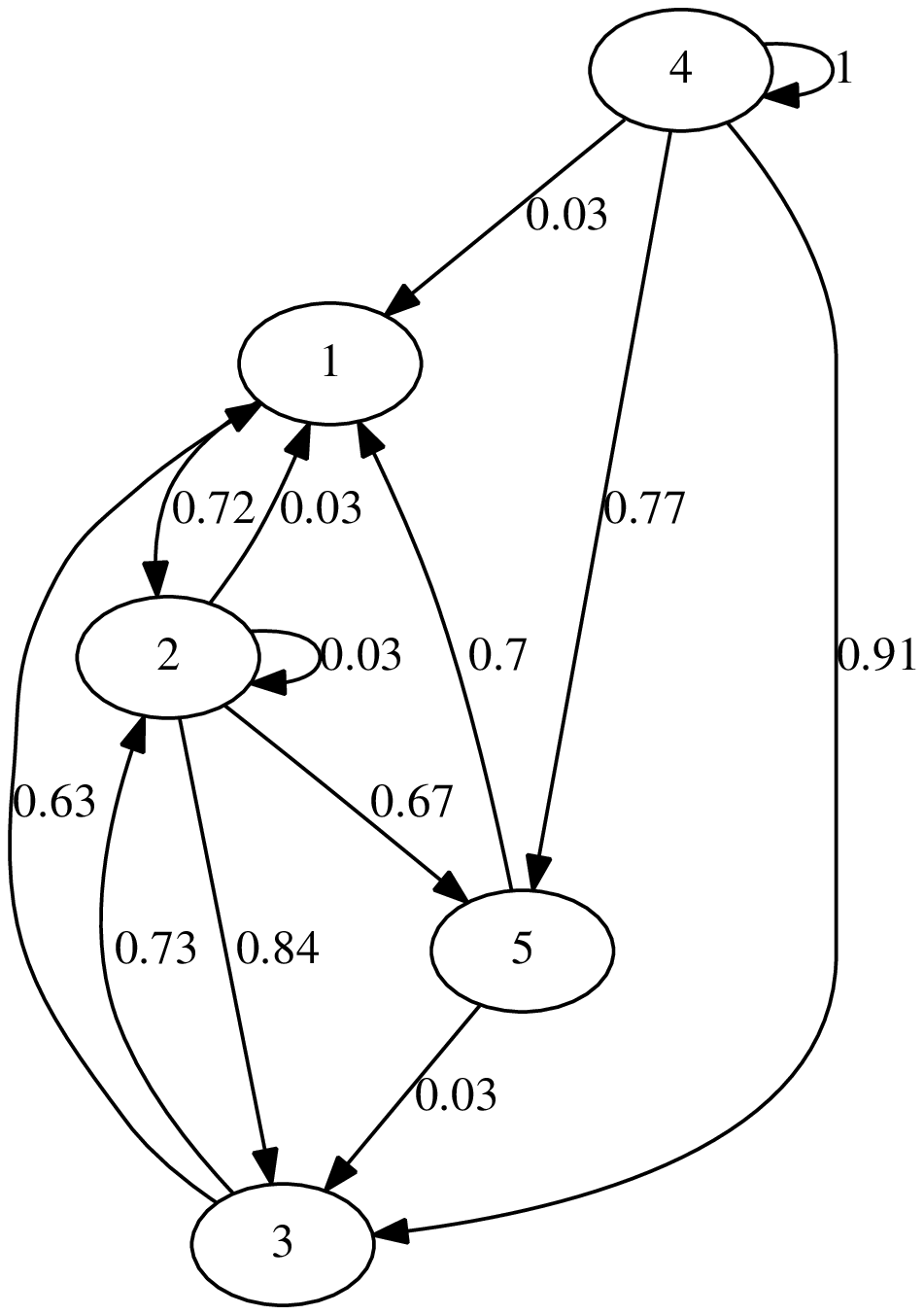}    } \hspace{-4pt}
  \subfigure[Mode 2]{\includegraphics[scale=0.29]{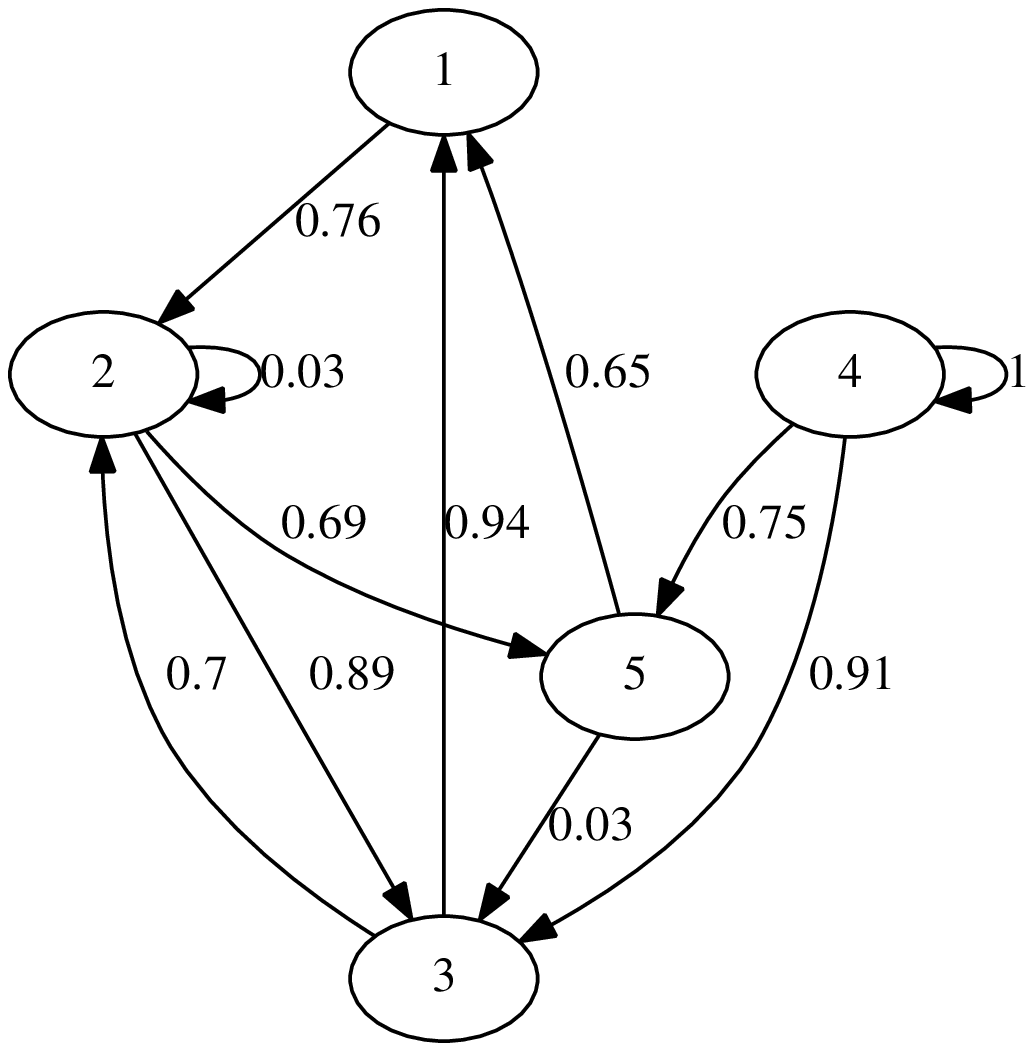}    } \hspace{-5pt}
  \subfigure[Mode 3]{\includegraphics[scale=0.29]{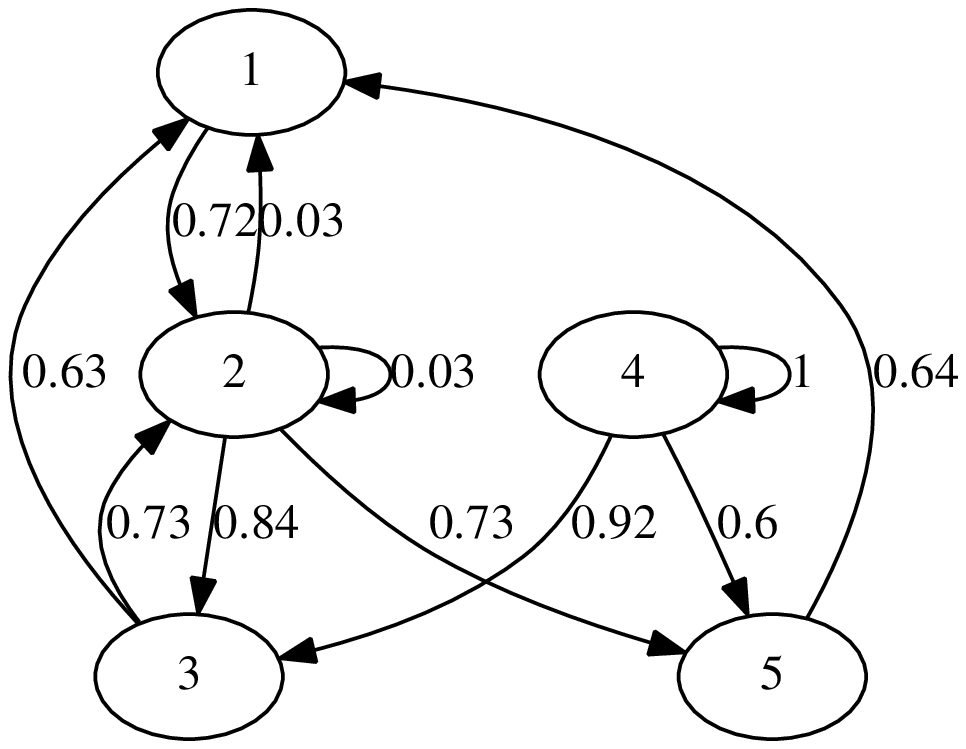}    } \hspace{-3pt}
  \subfigure[Mode 4]{\includegraphics[scale=0.29]{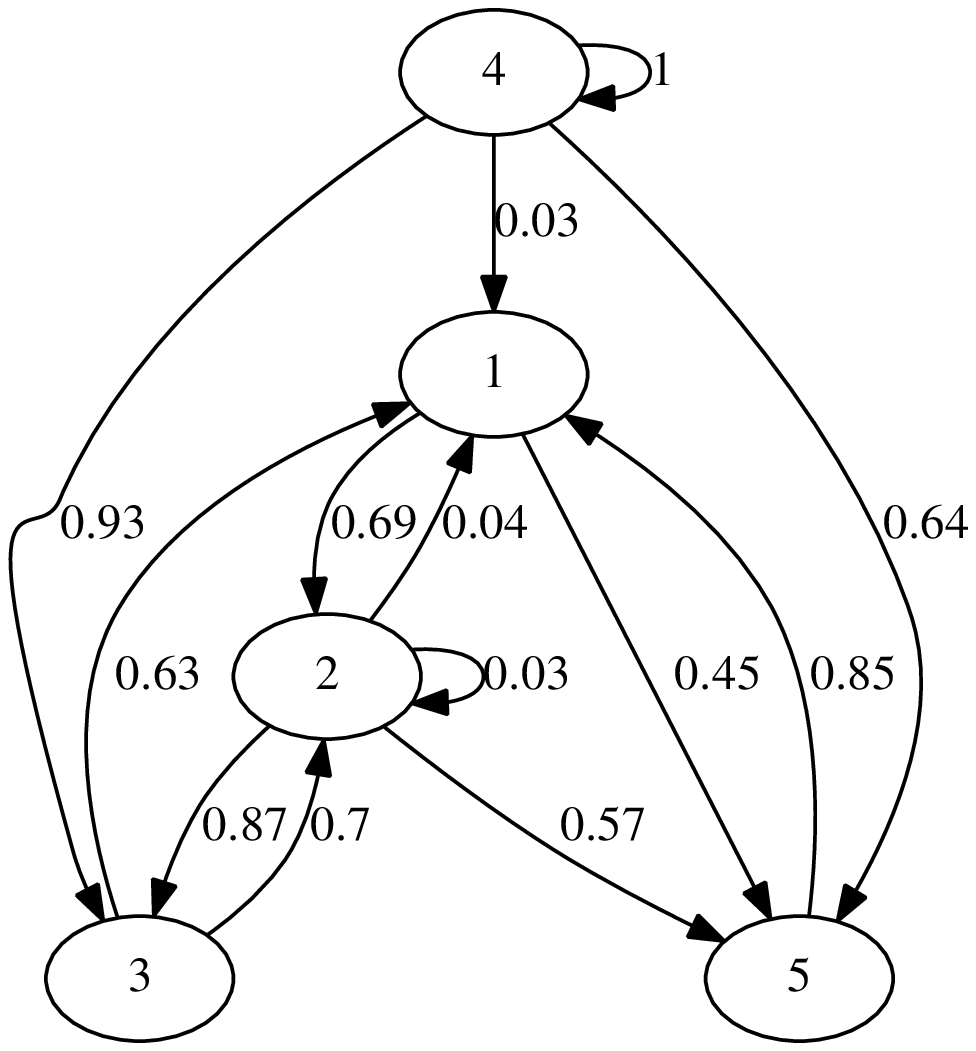}    } \hspace{-2pt}
  \subfigure[Mode 5]{\includegraphics[scale=0.29]{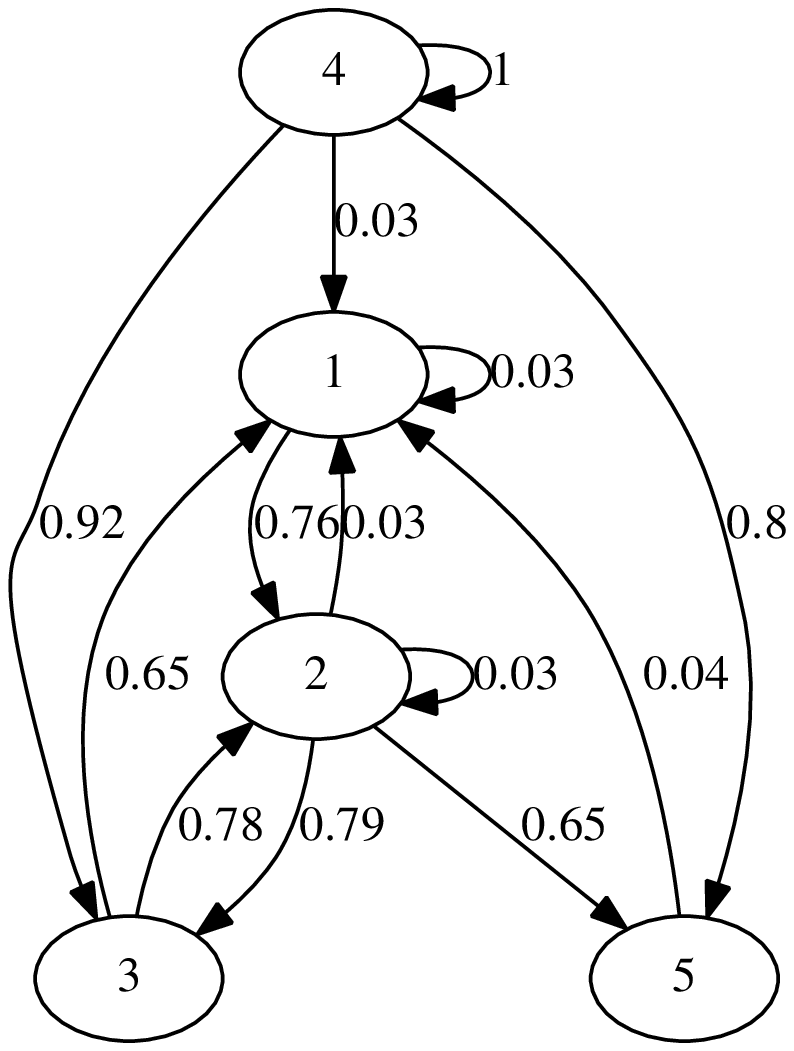}    } \hspace{-2pt}
  \subfigure[Mode 6]{\includegraphics[scale=0.29]{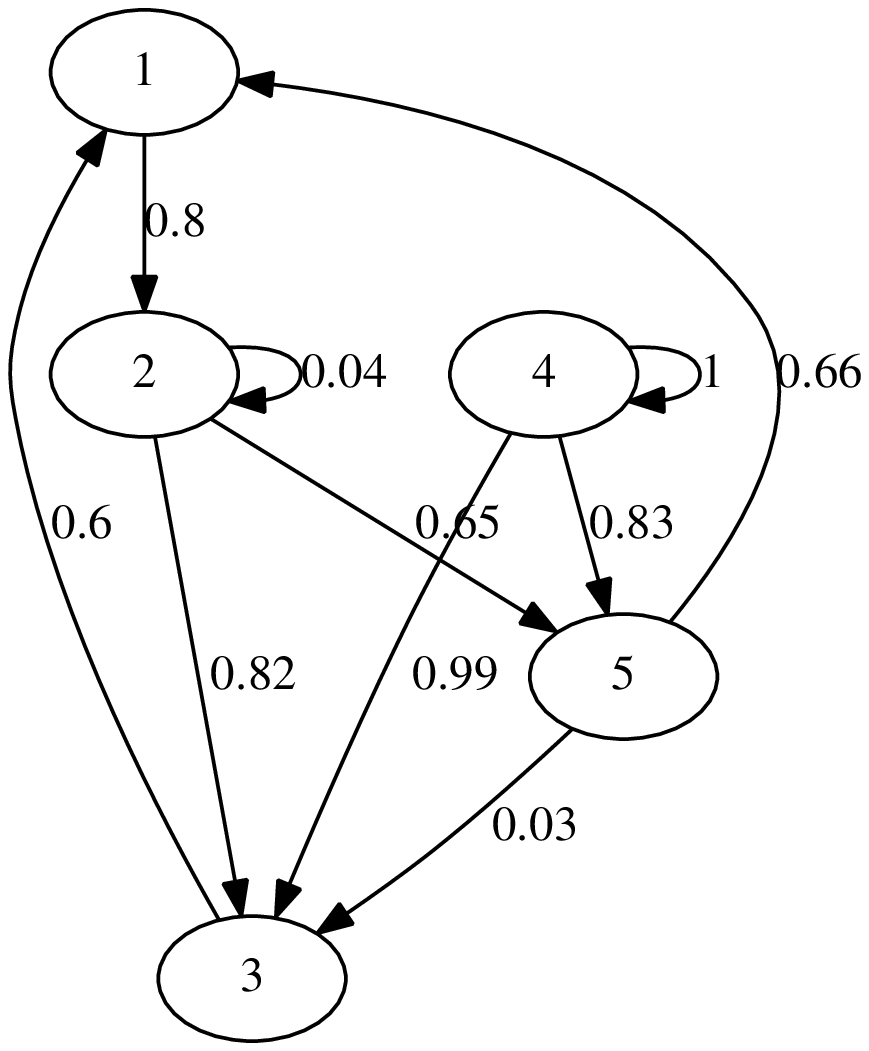}    } \vspace{-6pt}
  \caption{Graphical models defined to simulate pattern(s) anomaly. Six graphs are defined and treated as nominal operation modes in complex systems. Pattern failure is simulated by breaking specific patterns in the model (not shown).}
  \label{figMultimodes}
\end{figure*}

The pattern based root-cause analysis approaches proposed in Section \ref{secMethod} provide us the insight to identify the failed patterns in complex CPSs (corresponding to the cases that cyber-attacks may only influence interactions between specific sub-systems), and performance evaluation of $S^3$ and $A^3$ are carried out in Section \ref{subPatnRCA} including three synthetic datasets to simulate anomaly in pattern(s)/node(s), respectively. Using node inference algorithm in Section \ref{subsecNodeinf}, validation of the proposed approaches are illustrated in Section \ref{subNodeRCA}, with synthetic dataset and real data (TEP dataset ~\cite{russell2000data}).

\subsection{Pattern based root-cause analysis}
\label{subPatnRCA}
\subsubsection{Anomaly in pattern(s)}
\label{subsecAnoPat}
Anomaly in pattern(s) is defined as the change of one or more causal relationships. Anomaly translates to a changed/switched pattern in the context of STPN. A 5-node system is defined and shown in Fig.~\ref{figMultimodes} including six different nominal modes. Cycles ($1\to2\to5\to1$, $1\to2\to3\to1$, $2\to3\to2$) and loops ($4\to4$) are included in the models. The graphical models are applied to simulate multiple nominal modes. Anomalies are simulated by breaking specific patterns in the graph; 30 cases are formed including 5 cases in one failed pattern, 10 cases in two failed patterns, 10 cases in three failed patterns, and 5 cases in four failed patterns. Multivariate time series data (denoted as \texttt{dataset1}) are generated using VAR process that follows the causality definition in the graphical models. The VAR process is widely used in economics and other sciences for its flexibility and simplicity for generating multivariate time series data \citep{goebel2003investigating}. The basic VAR model for $Y(t)={y_{i,t}, i=(1,2,\cdots,f),\ t \in \mathbb{N}}$ is defined as
\begin{equation}
y_{i,t}=\sum_{k=1}^{p}{\sum_{j=1}^{n}{A_{i,j}^{k}y_{j,t-k}+\mu_{t}}}, \quad j=(1,2,\cdots,f)
\label{eqVARmodel}
\end{equation}
where $p$ is time lag order, $A_{i,j}$ is the coefficient of the influence on $i$-th time series caused by $j$th time series, and $\mu_{t}$ is an error process with the assumption of white noise with zero mean, that is $E(\mu_{t})=0$, the covariance matrix, $E(\mu_{t}\mu_{t}^{\prime})=\Sigma_{\mu}$ is time invariant.

\textbf{STPN parameters:} To generate the inputs for RBM and DNN, MEP is applied to partition the time-series into 9 bins and the length of the short-time symbol sequence is 1200.

\textbf{Performance Evaluation:} Root-cause identification performances of $S^3$ and $A^3$ methods are evaluated using \texttt{dataset1}.
The accuracy metric used here is a pattern-by-pattern comparison (both anomalous and nominal) between the prediction and the ground truth labels.
A successful match occurs if the prediction states that a particular node is not faulty, and if the ground truth is, in fact, not faulty; the same goes for anomalies. Otherwise, the prediction is considered an error. Formally, we define the accuracy metric $\alpha_1$ as:
\begin{equation}
\alpha_1 = \frac{\sum_{j=1}^{n^{2}} \sum_{i=1}^m \boldsymbol{\chi_1}(T_{ij}=P_{ij})}{m n^{2}}
\label{eqalpha1}
\end{equation}
where $T_{ij}$ denotes the ground truth state (nominal/anomalous) of the $j^{th}$ pattern of the $i^{th}$ test sample. Similarly, $P_{ij}$ is the corresponding predicted state using the root-cause analysis approach. $\boldsymbol{\chi_1(\cdot)}$ is the indicator function that returns the value 1 if the expression within the function is true. In the denominator, $m$ is the number of test samples and $n^{2}$ is the number of patterns.
With the above metric, results of $S^3$ method and $A^3$ method are listed in Table~\ref{tabAccu}. Precision and F-measure are also evaluated with the definitions as follows~\cite{fawcett2006introduction}.
\[\text{Recall}=\frac{TP}{TP+FN}\text{, }
\text{Precision}=\frac{TP}{TP+FP}\text{,}\]
\[\text{F-measure}=\frac{2}{1/\text{Recall}+1/\text{Precision}}\text{,}\]
where $TP$ is the number of true positives (the correctly detected anomalous patterns), $FN$ is the number of false negatives \color{black}(the undetected anomalous patterns), and $FP$ is the number of false positives (the detected anomalous patterns which are indeed nominal).

\begin{table}[htbp]
\caption{Root-cause analysis results in $S^3$ method and $A^3$ method with synthetic data.}
\centering
\label{tabAccu}
\begin{tabular}{c c c c c c c}
\hline
  \multirow{2}*{\small{Method}} &Training/Testing & Accuracy & Recall/Precision\\
   &   samples & $\alpha_1 (\%)$ & /F-measure (\%)\\
  \hline
  $S^3$ &11,400/57,000 &97.04 &99.40/97.10/98.24\\ 
  $A^3$ &296,400/57,000 &98.66 &90.46/95.95/93.12\\
\hline
\end{tabular}
\end{table}

High \emph{accuracy} is obtained for both $S^3$ and $A^3$ method. While training time is much less for $S^3$, the inference time in root-cause analysis for $S^3$ is much more than that of $A^3$, as $S^3$ depends on sequential searching. Note, the classification formulation in $A^3$ aims to achieve the exact set of anomalous nodes. On the other hand, the $S^3$ method is an approximate method that sequentially identifies anomalous patterns and hence, the stopping criteria would be critical. The observation that the performance of $S^3$ is quite comparable to that of $A^3$ suggests a reasonable choice of the stopping criteria.

\subsubsection{Anomaly in node(s)}
\label{subsecAnoNod}
\begin{figure*}[htbp]
  \centering
  \subfigure[Fault node 1]{\includegraphics[trim={0 0 0 20},scale=0.3]{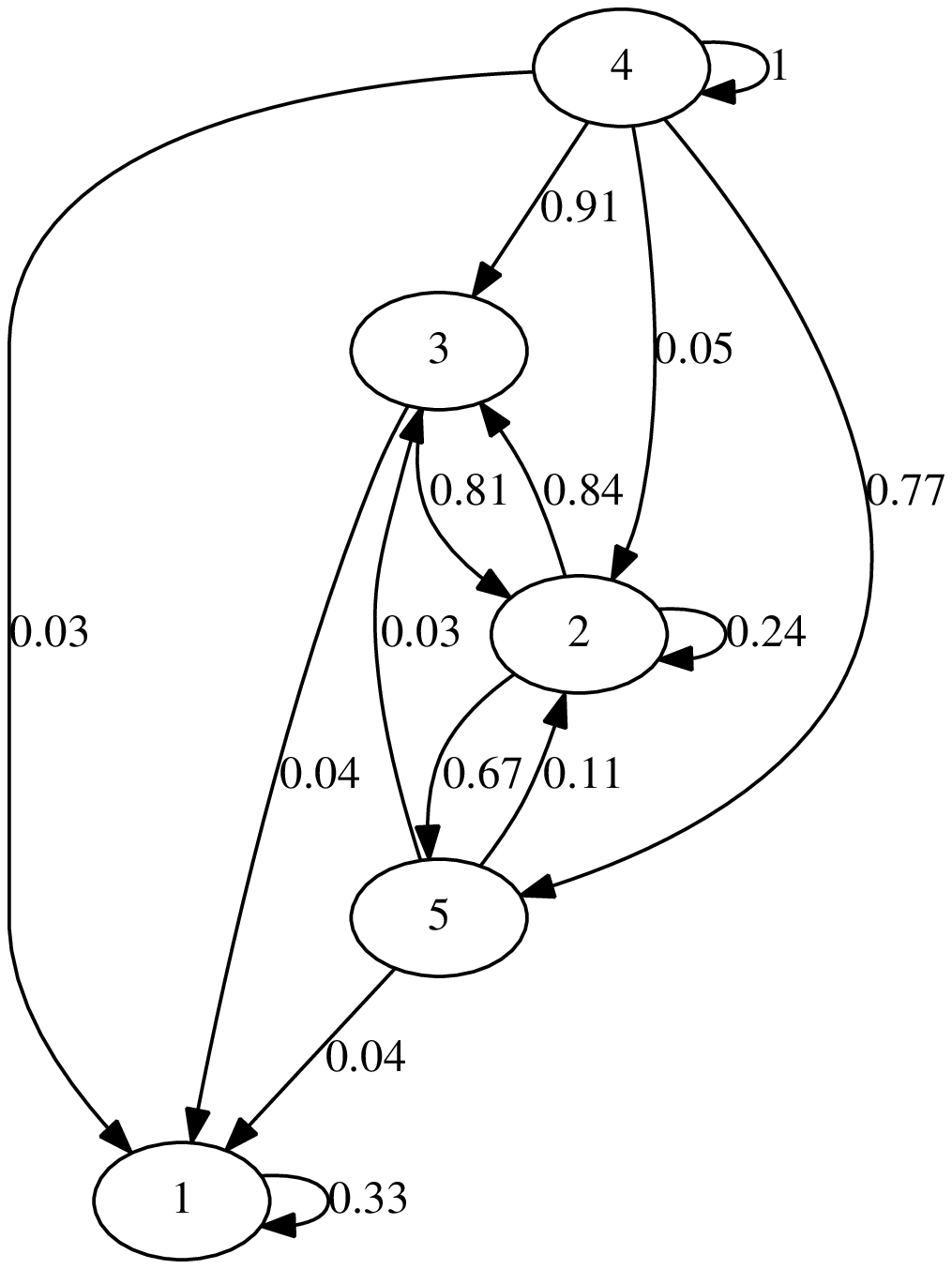}    } \hspace{-1pt}
  \subfigure[Fault node 2]{\includegraphics[scale=0.3]{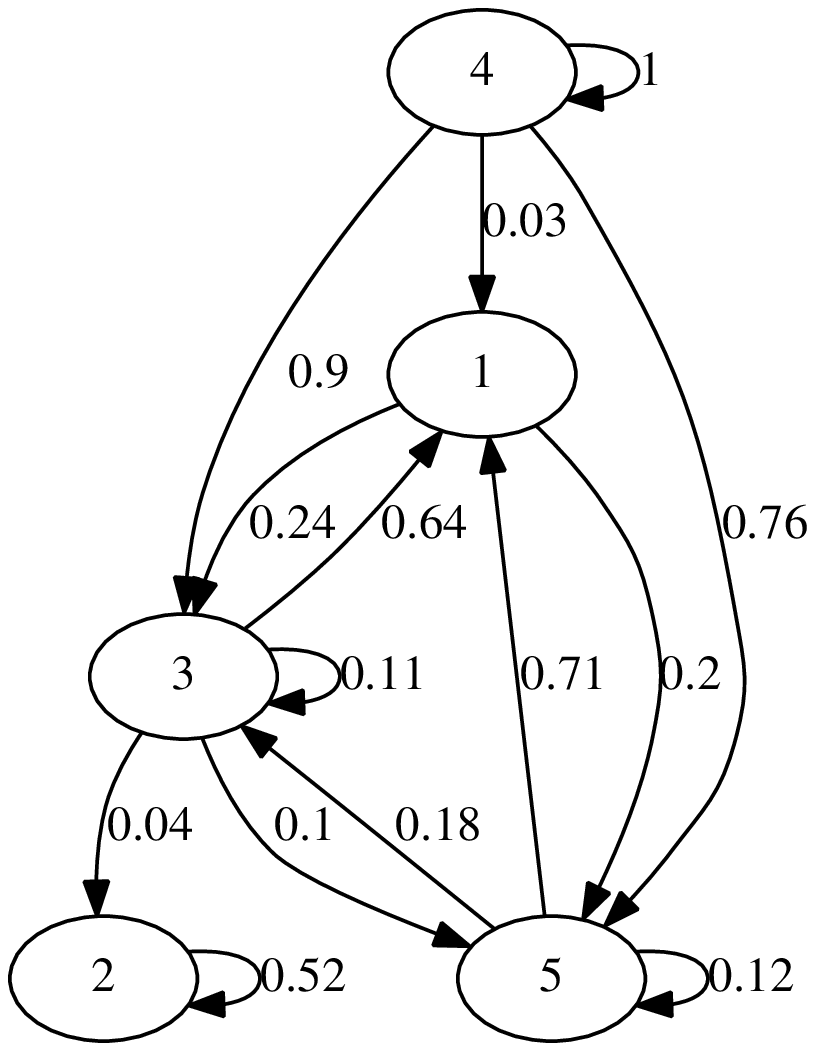}    } \hspace{8pt}
  \subfigure[Fault node 3]{\includegraphics[scale=0.3]{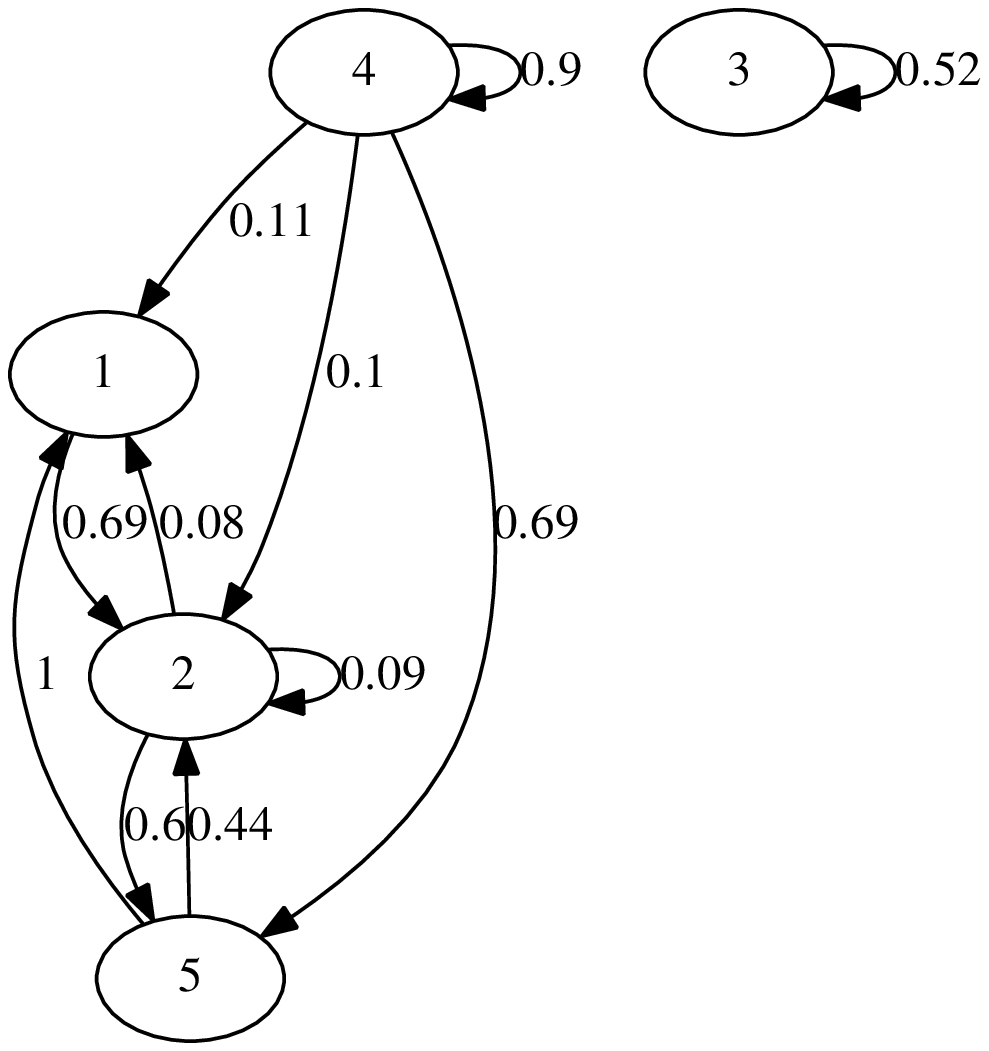}    } \hspace{-1pt}
  \subfigure[Fault node 4]{\includegraphics[scale=0.3]{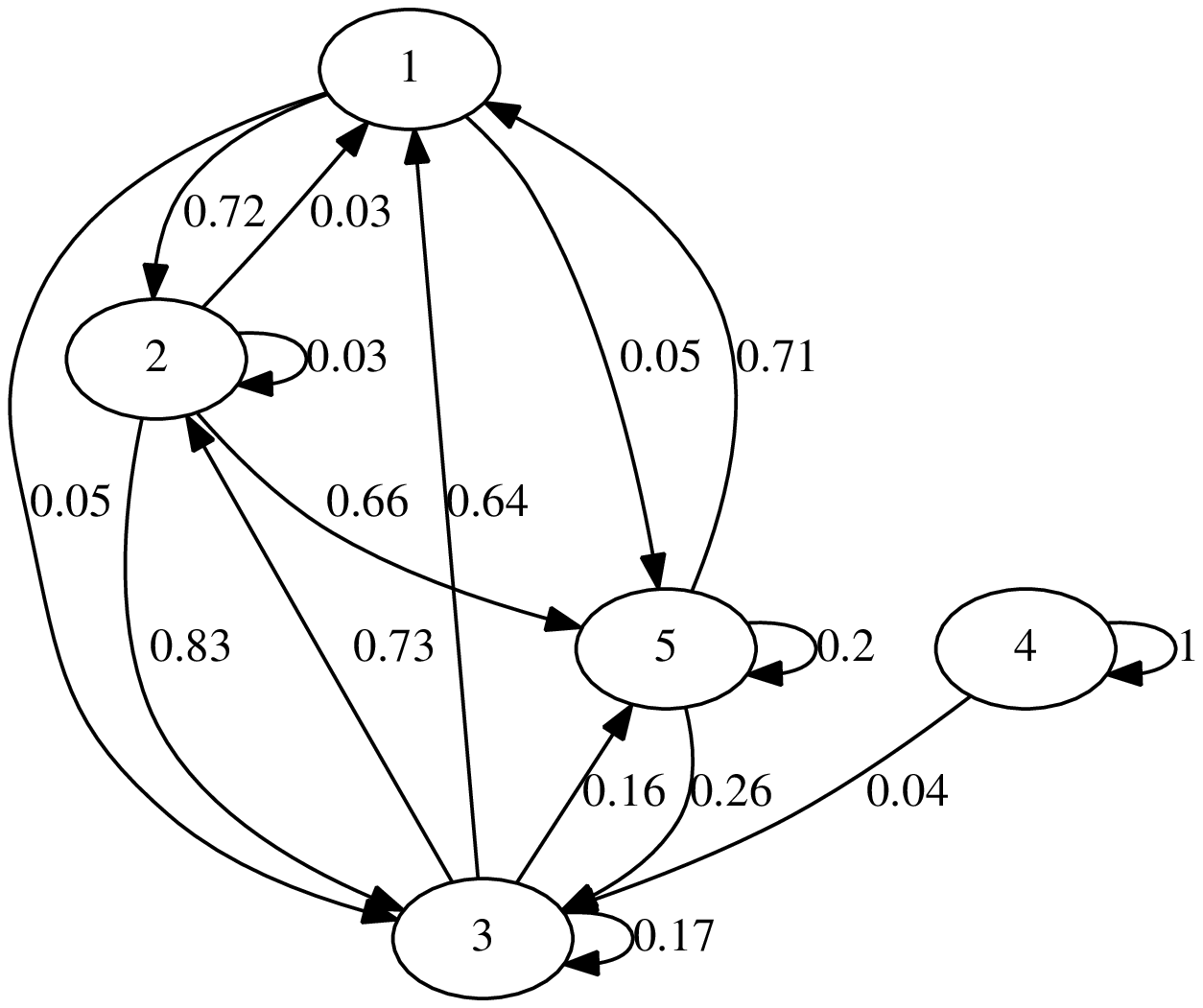}    } \hspace{1pt}
  \subfigure[Fault node 5]{\includegraphics[scale=0.3]{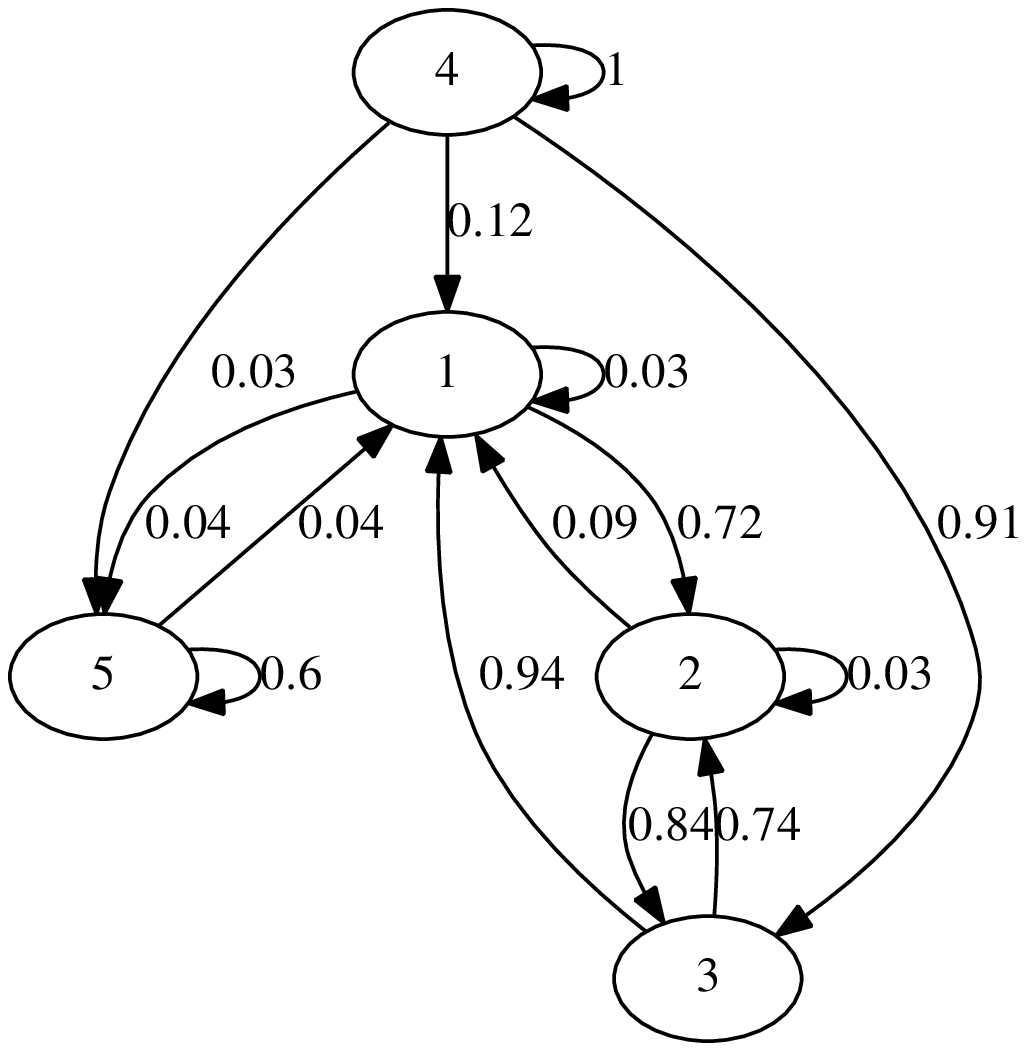}    } \vspace{-6pt}
  \caption{Anomalous conditions with fault node, causality is discovered by VAR model. Compared with the nominal mode in Fig.~\ref{figMultimodes} (a), the fault node breaks most of the causality from and to this node. Causality is normalized by the maximal value among all of the patterns, and causality smaller than 0.03 is not shown.} 
  \label{figNodefailure}
\end{figure*}

\textbf{Dataset:} Anomaly in node(s) occurs when one node or multiple nodes fail in the system. For a cyber-physical system, they may be caused by sensor fault or component degradation. The graphical model defined in Fig.~\ref{figMultimodes} (a) is applied for generating the nominal data using the VAR process. Anomaly data are simulated by introducing time delay in a specific node. The time delay will break most of the causal relationship to and from this node (except possibly the self loop, i.e., AP of the failed node). For validation and comparison, graphs are recovered with VAR process in cases of faulty nodes as shown in Fig.~\ref{figNodefailure}). The figure also shows that there are some variations in causality between nominal nodes which suggests that causality discovery is more difficult with the existence of cycles and loops. The generated dataset is denoted as \texttt{dataset2}. For scalability analysis, a 30-node system is further defined using VAR process, and the same method is applied to simulate the anomaly data, noted as \texttt{dataset3}.

\textbf{Performance Evaluation:} This work is aimed at discovering failed patterns instead of recovering underlying graph (which is very difficult in graphs with cycles and loops~\cite{richardson1996discovery,lacerda2008discovering}). The discovered anomalous patterns can then be used for diagnosing the fault node. For instance, a failed pattern $N_{i}\to N_{j}$ discovered by root-cause analysis can be caused by the fault node $i$ or $j$. However, if multiple failed patterns are related to the node $i$, then this node can be deemed anomalous. In this regard, it is important to learn the impact of one pattern on a detected anomaly compared to another. This can facilitate a ranking of the failed patterns and enable a robust isolation of an anomalous node, which is the motivation of the node inference algorithm.

For comparison, we use VAR-based graph recovery method that is widely applied in economics and other sciences, and efficient in discovering Granger causality \cite{goebel2003investigating}. Note, the test dataset itself is synthetically generated using a VAR model with a specific time delay. Hence, the causality in such a multivariate time series is supposed to be well captured by VAR-based method. The details of the VAR-based root-cause analysis strategy is explained below.

With the given time series, a VAR model (i.e., the coefficients $A_{i,j}$ in Eq. \ref{eqVARmodel}) can be learned using standard algorithm~\cite{goebel2003investigating}. The differences in coefficients  between the nominal and anomalous models are subsequently used to find out the root causes. The pattern is deemed to have failed when $\delta A_{i,j}>\eta \cdot \max\{\delta A_{i,j}\}$ where $\delta A_{i,j}=|A_{i,j}^{ano}-A_{i,j}^{nom}|$, $\eta =0.4$.

\begin{figure*}[htbp]
  \centering
  \subfigure[Fault in Node 1]{\includegraphics[width=0.45\textwidth,clip=True]{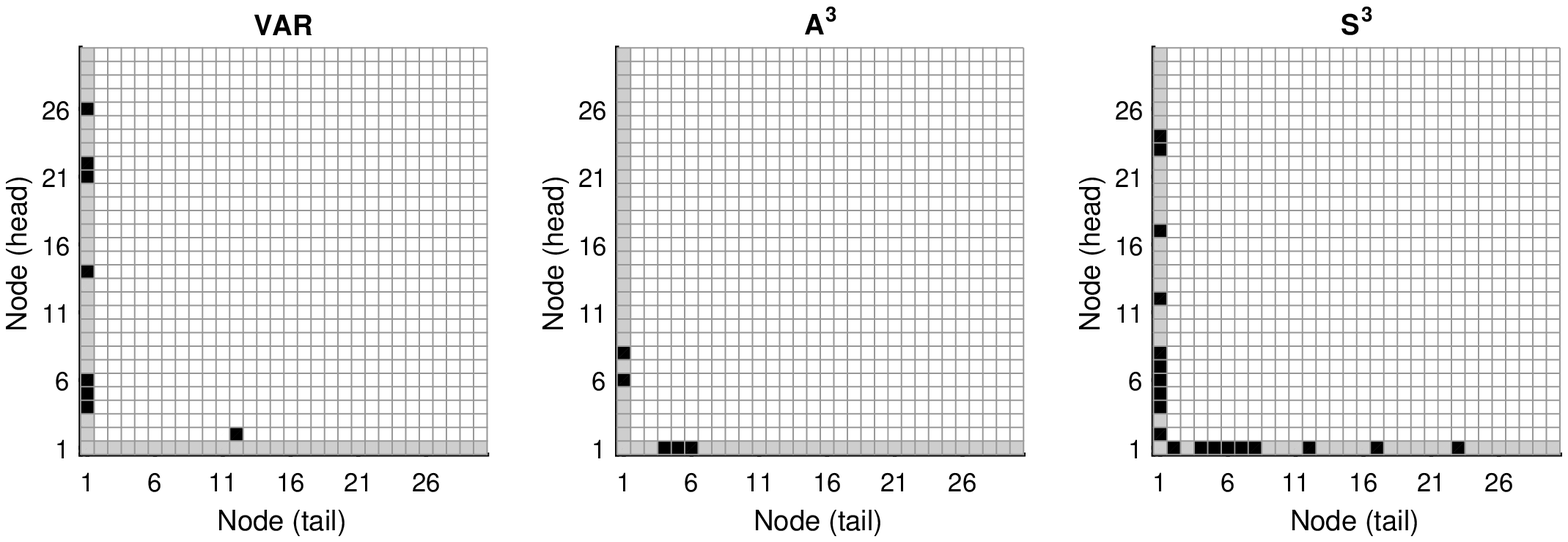}} \hspace{10pt}
  \subfigure[Fault in Node 12]{\includegraphics[width=0.45\textwidth,clip=True]{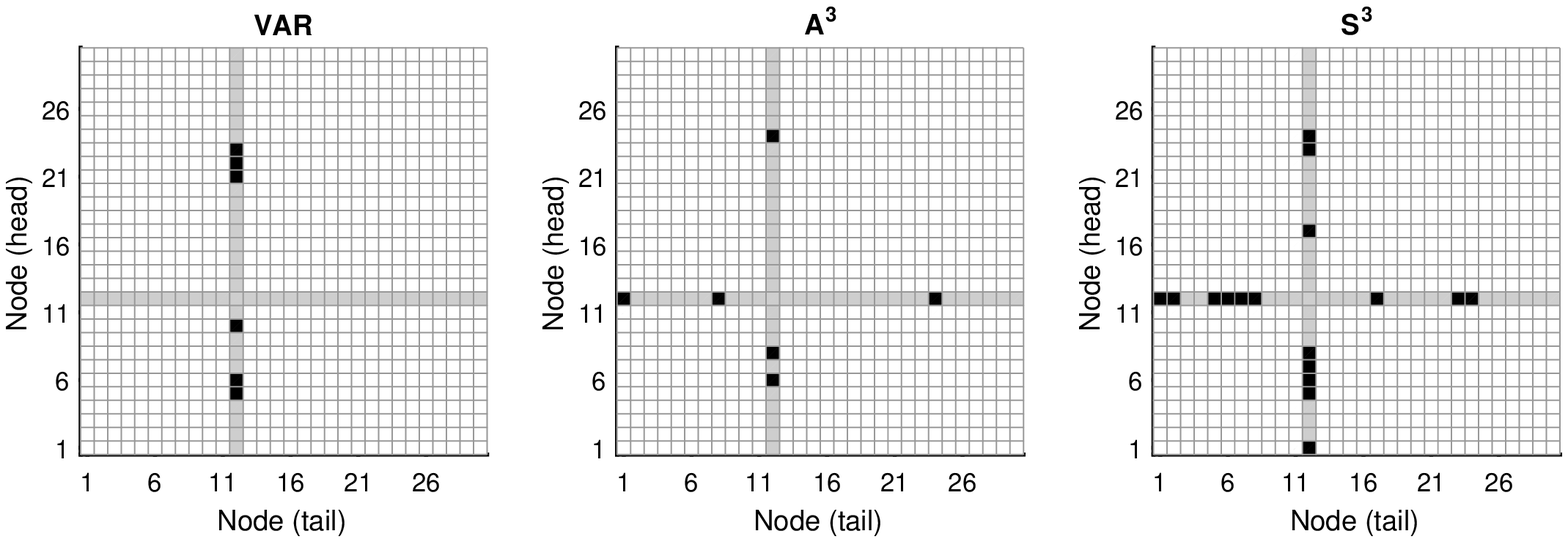}} \\ 
  \subfigure[Fault in Node 25]{\includegraphics[width=0.45\textwidth,clip=True]{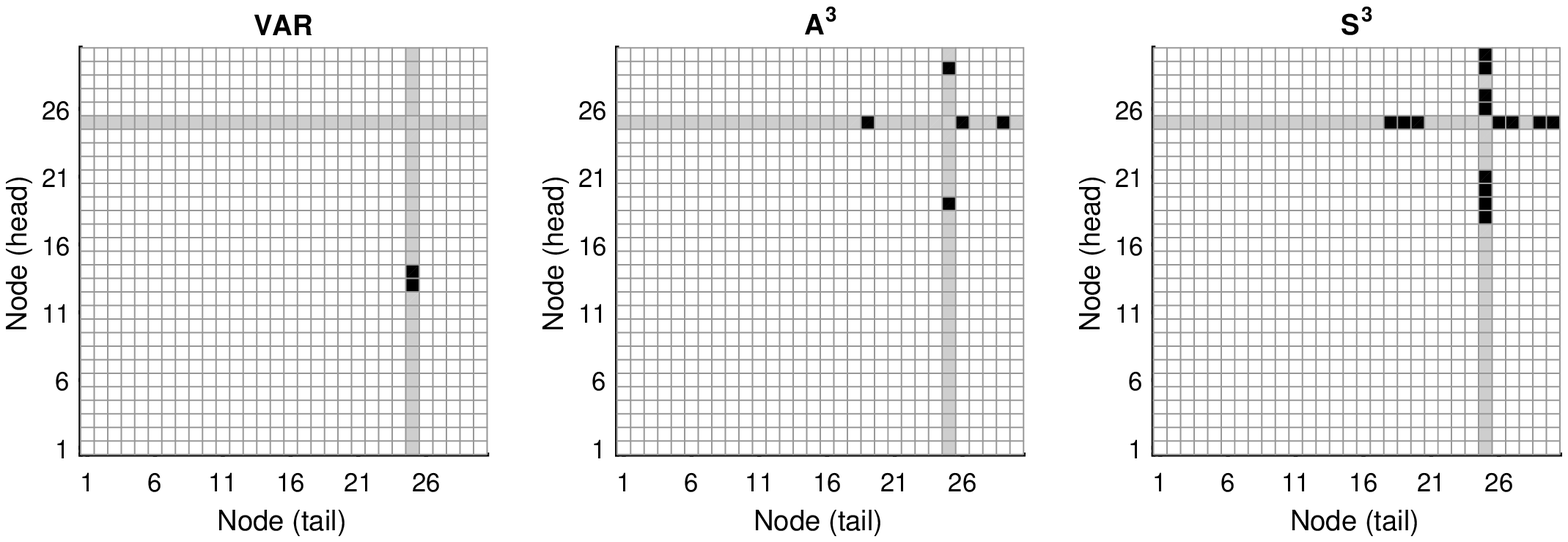}} \hspace{10pt}
  \subfigure[Fault in Node 30]{\includegraphics[width=0.45\textwidth,clip=True]{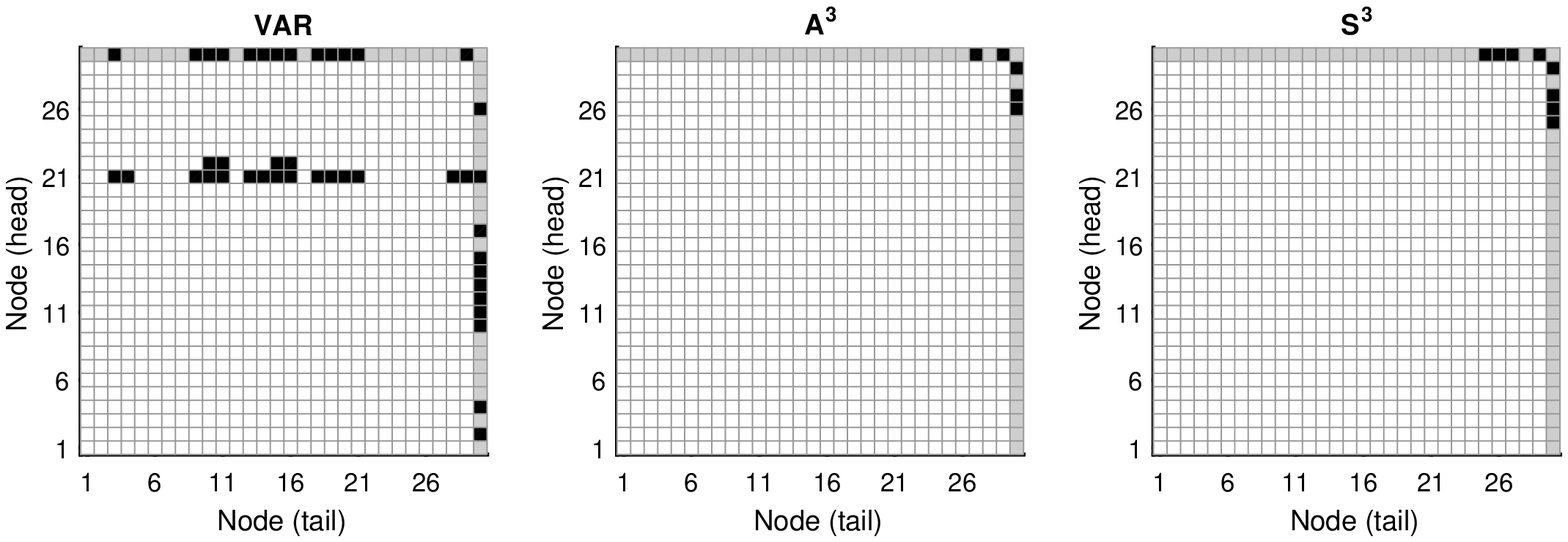}}\\ 
  \caption{Comparisons of artificial anomaly association ($A^3$), sequential state switching ($S^3$) method and vector autoregressive (VAR) method using \texttt{dataset3}. The patterns represented with boxes are from the tail node in x-axis to the head node in y-axis, and the discovered root causes are marked in black. The boxes in gray are corresponding to the node with fault injected. The patterns (in black boxes) in the gray zone are the patterns correctly discovered.  } 
  \label{figRCA30node}
\end{figure*}

The results of $S^3$ and VAR using \texttt{dataset3} are shown in Fig.~\ref{figRCA30node}. In panel (a), all of the changed patterns discovered by $A^3$ and $S^3$ can be attributed to node 1 (shown by the black boxes in column and row 1). Therefore, node 1 is considered as faulty by the $A^3$ and $S^3$ method. On the other hand, VAR incorrectly discovers a significant change in the pattern $N_{12}\to N_{2}$ but not the patterns originating from $N_1$, and this will be interpreted as failed nodes $N_{12}$ and $N_{2}$, we note this as an error. In general, although $A^3$, $S^3$ and VAR can discover the fault node, VAR produces more false alarms. In cases (b) and (c), $A^3$, $S^3$ find out more correct patterns than VAR, and the discovery of more changed patterns can contribute to a stronger evidence that node $12$ and $25$ are faulty, respectively. In general, VAR discovers more patterns than $A^3$, $S^3$, while quite a lot of them are attributed to false (as shown in panel (d)), and this causes more false positives in VAR. For $A^3$ and $S^3$, $A^3$ discovers fewer patterns, while most of them are correct, and this means higher precision in $A^3$. $S^3$ finds out more patterns than $A^3$, indicating higher recall.

Note that accuracy metrics of recall and precision cannot be directly computed in this dataset as the injected fault may not change/flip all of the patterns connected to the faulty node (gray zone in Fig.~\ref{figRCA30node}), depending on the original graphical model in nominal condition, while the accurate graph for nominal condition in complex CPSs is difficult to be discovered~\cite{richardson1996discovery,lacerda2008discovering}.

In real applications, when more anomalous patterns are discovered incorrectly, more effort will be needed to analyze the failed patterns closely and determine the root-cause node. This will lead to more financial expenditures and time investment in finding the true root-cause. With this motivation, an error metric $\epsilon$ is defined by computing the ratio of incorrectly discovered anomalous patterns $|\{\Lambda^{\epsilon}\}|$ to all discovered anomalous patterns $|\{\Lambda^{ano}\}|$, i.e., $\epsilon={|\{\Lambda^{\epsilon}\}|}/{|\{\Lambda^{ano}\}|}$. The results using \texttt{dataset2} and \texttt{dataset3} are listed in Table~\ref{tabAccu2}. Note that the error metric $\epsilon$ is defined to address the following key issue: we do not have the ground truth for the computation of the standard metrics such as precision and recall. However, if we compare the definition of error metric and precision, we obtain $\epsilon = 1- precision$ if the ground truth is known. However, for \texttt{dataset2} and \texttt{dataset3}, we inject the faulty node by adding time delay. The time delay breaks most causal relationships to and from this node, but not all of them are broken. In this context, we cannot compute the precision and recall for the two datasets.

\begin{table}[htbp]
\caption{Comparison of root-cause analysis results with VAR, $A^3$, and $S^3$.}
\centering
\label{tabAccu2}
\begin{tabular}{c c c c c c c}
\hline
  \multirow{2}*{Method} & \multicolumn{3}{c}{Dataset 2 (5 nodes)}  & \multicolumn{3}{c}{Dataset 3 (30 nodes)} \\
    &$|\{\Lambda^{ano}\}|$ &$|\{\Lambda^{\epsilon}\}|$ &$\epsilon$ (\%)  &$|\{\Lambda^{ano}\}|$ &$|\{\Lambda^{\epsilon}\}|$ &$\epsilon$ (\%) \\
  \hline
  VAR  &20 &4 &20.0   &521 &113 &21.7\\
  $A^3$  &24 &2 &8.3  &105 &1 &0.95\\
  $S^3$  &13 &1 &7.7  &653 &0 &0\\
\hline
\end{tabular}
\end{table}

While it should be noted that $A^3$, $S^3$ and VAR can mostly discover the fault node correctly in both the data sets, the error ratios for $A^3$ and $S^3$ method are much lower than that for VAR (i.e., lower false alarm). As the scale of the system increases, the number of discovered anomalous patterns by $S^3$ becomes more than that of by VAR, while the error ratio becomes significantly less than that of VAR. Therefore, $S^3$ method is both \emph{scalable} as well as demonstrates better accuracy. Note that for comparisons between $A^3$, $S^3$ and VAR, only one nominal mode is considered in Table~\ref{tabAccu2} as VAR is not directly applicable in cases with multiple nominal modes. $A^3$ and $S^3$ methods can handle multiple nominal modes and it has been validated in Section \ref{subsecAnoPat}.

It should also be noted that the metrics listed in Table \ref{tabAccu} are not applicable in this dataset, as of the anomaly injection approach used in this work. Introducing time delays at a node level may influence most of the existing interactions between the node and the other nodes (noted as flipped pattern in this work). However, the pattern will not be changed or flipped if the interaction in nominal condition is weak or does not exist. As the approaches has been validated via \texttt{dataset1}, the results of \texttt{dataset2} (Table~\ref{tabAccu2}) and \texttt{dataset3} (Fig.~\ref{figRCA30node}, Table~\ref{tabAccu2}) intend to show that the proposed approaches are also capable of node inference via pattern based root-cause analysis.

\begin{rem}
The root-cause analysis algorithms proposed in this work are fundamentally pattern-based as opposed to being node-based. The motivation for such methods comes from real cyber-physical systems, where cyber-attacks may only compromise interactions among sub-systems (i.e., relational patterns) without directly affecting sub-systems (i.e., the nodes). In contrast, the majority of the existing methods mostly focus on node-based anomalies to the best of our knowledge. For example, in~\cite{qiu2012granger}, root-cause analysis was performed using Granger Graphical Model with neighborhood similarity (GGM-N) and Granger Graphical Model with coefficient similarity (GGM-C). Using node inference algorithm, the fault nodes can be obtained based on failed patterns, and case studies are illustrated in the next section.
\end{rem}
\subsection{Root-cause analysis with node inference}
\label{subNodeRCA}
Synthetic data sets and a real data set (T...) are applied in this section for inferring the failed node using the proposed approach as well as performance comparisons with VAR, GGM-N and GGM-C.

\subsubsection{Node based root-cause analysis with synthetic data}
\label{secsupNodeRCA}

With root-cause results in Section \ref{subsecAnoNod}, node inference is implemented using Algorithm 2, the results of \texttt{dataset3} are listed in Table \ref{tabNodeAccu}. Note that node inference for VAR is also implemented using Algorithm 2, with the weights set as 1 for each failed pattern. The performance in terms of recall, precision and F-measure of $A^{3}$ and $S^{3}$ are compared with VAR, GGM-N, and GGM-C (algorithms of GGM-N and GGM-C are in \cite{qiu2012granger}, parameters are as follows, $\lambda=30$, $45$; $\lambda_{1}=10$, $10$; $\lambda_{2}=10$, $10$ for GGM-N and GGM-C, respectively).

\begin{table}[htbp]
\caption{Node inference results in VAR, GGM-N, GGM-C, $A^3$ and $S^3$ methods with synthetic data.}\vspace{3pt}
\centering
\label{tabNodeAccu}
\begin{tabular}{c c c c c c c}
\hline
  Approach & Recall (\%) & Precision (\%) &F-measure (\%)\\
  \hline
  VAR &100 &50.8 &67.4 \\
  GGM-N &83.3 &100 &90.9 \\
  GGM-C &80.0 &96.0 &87.3 \\
  $A^3$ &96.7 &90.6 &93.6 \\
  $S^3$ &100 &100  &100 \\
\hline
\end{tabular}
\end{table}

\subsubsection{Node inference with real dataset -- Tennessee Eastman process (TEP)}
\label{subsubsecTEP}
TEP data is based on a realistic simulation program of a chemical plant from the Eastman Chemical Company, USA, and it has been widely used for process monitoring community as a source of data for comparing various approaches, and a benchmark for control and monitoring studies~\cite{russell2000data,yin2012comparison,downs1993plant}. The process consists of five major units: reactor, condenser, compressor, separator, and stripper, with 53 variables simulated including 41 measured and 12 manipulated (the agitation speed is not included in TEP dataset as it is not manipulated). 21 faults are simulated in TEP program, while 8 of them are applied in this work, as the root causes of these faults are intuitively explained in \cite{russell2000data}.

\begin{table}[htbp]
\caption{Methods for node inference with TEP dataset.}
\centering
\label{tabMethodsList}
\begin{threeparttable}
\begin{tabular}{l l l}
\hline
  Method & Description \\
  \hline
  PCA-CONT & PCA, total contribution based \\
  PCA-RES & PCA, residual based \\
  DPCA-CONT & DPCA projection, total contribution based\\
  DPCA-RES &DPCA projection, residual based\\
  CVA-CONT &CVA, total contribution based \\
  CVA-RES & CVA projection, residual based\\
  GGM-N & GGM with neighborhood similarity\\
  GGM-C & GGM with coefficient similarity\\
  $A^{3}$ & Artificial anomaly association\\
  $S^3$ &Sequential state switching\\
\hline
\end{tabular}
  \begin{tablenotes}
    \item[] PCA--Principal Component Analysis, DPCA--Dynamic PCA, CVA--Canonical Variate Analysis, GGM--Granger graphical model.
  \end{tablenotes}
\end{threeparttable}
\end{table}

\begin{table}[htbp]
\centering
\caption{The rankings of node inference results on TEP dataset}
\label{tabTEPrank}
\begin{threeparttable}
\begin{tabular}{c c c c c c c c c c c c}
\hline
  \multicolumn{2}{c}{Fault}  & \multicolumn{2}{c}{GGM$^{b}$}   &\multicolumn{2}{c}{PCA$^{a}$} &\multicolumn{2}{c}{DPCA$^{a}$} & \multicolumn{2}{c}{CVA$^{a}$}  &\multirow{2}*{$A^{3}$} &\multirow{2}*{$S^{3}$} \\
   & &\scriptsize{-N} &\scriptsize{-C} &\scriptsize{-CONT} &\scriptsize{-RES} &\scriptsize{-CONT} &\scriptsize{-RES} &\scriptsize{-CONT} &\scriptsize{-RES}   & & \\ \hline
	\multirow{2}*{2} &0-15h	&	-	&	-	&	2	&	4	&	2	&	5	&	10	&	2	&	30	&	5	\\	
	&15-40h	&	-	&	-	&	2	&	5	&	3	&	12	&	10	&	4	&	31	&	9	\\	\hline
	\multirow{2}*{4} &0-15h	&	2	&	2	&	1	&	1	&	1	&	1	&	10	&	1	&	1	&	1	\\	
	&15-40h	&	1	&	1	&	1	&	1	&	1	&	1	&	11	&	1	&	2	&	1	\\	\hline
	\multirow{2}*{5} &0-15h	&	2	&	2	&	12	&	21	&	11	&	8	&	15	&	17	&	21	&	2	\\	
	&15-40h	&	4	&	4	&	9	&	35	&	14	&	30	&	16	&	16	&	18	&	2	\\	\hline
	\multirow{2}*{6} &0-15h	&	-	&	-	&	1	&	6	&	3	&	2	&	6	&	6	&	8	&	2	\\	
	&15-40h	&	-	&	-	&	7	&	51	&	11	&	45	&	1	&	3	&	2	&	21	\\	\hline
	\multirow{2}*{11} &0-15h	&	2	&	2	&	1	&	1	&	1	&	1	&	10	&	1	&	1	&	1	\\	
	&15-40h	&	2	&	2	&	1	&	1	&	1	&	1	&	13	&	1	&	1	&	1	\\	\hline
	\multirow{2}*{12} &0-15h	&	-	&	-	&	1	&	6	&	1	&	3	&	10	&	14	&	4	&	1	\\	
	&15-40h	&	-	&	-	&	10	&	21	&	4	&	36	&	17	&	26	&	3	&	1	\\	\hline
	\multirow{2}*{14} &0-15h	&	-	&	-	&	2	&	2	&	1	&	2	&	11	&	1	&	4	&	1	\\	
	&15-40h	&	-	&	-	&	2	&	2	&	1	&	2	&	11	&	1	&	13	&	1	\\	\hline
	\multirow{2}*{21} &0-15h	&	-	&	-	&	52	&	40	&	48	&	48	&	52	&	52	&	4	&	1	\\	
	&15-40h	&	-	&	-	&	52	&	48	&	52	&	52	&	52	&	50	&	3	&	1	\\	\hline
\end{tabular}
  \begin{tablenotes}
     \item[a] The results of PCA, DPCA, and CVA are from \cite{russell2000data}, where 0-5h results are used (as shown in  \cite{russell2000data}. As time progresses, the accuracy decreases; the best results are listed here. These are rankings of the potential root-cause variables which are most closely related to the anomaly. The lower the number, the better the performance.
     \item[b] The results of GGM-N and GGM-C are from \cite{qiu2012granger}.
  \end{tablenotes}
\end{threeparttable}
\end{table}

For comparison, the methods in the state-of-the-art are used as listed in Table~\ref{tabMethodsList}. The root-cause results are shown in Table \ref{tabTEPrank}, which are essentially root cause potential rankings of the variable (most closely related to the anomaly), so the lower number indicates better performance.

According to \cite{russell2000data, qiu2012granger}, faults 4, 5, and 11 are the most difficult ones to be detected. However, the rankings obtained in \cite{russell2000data} show that faults 21, 5, 12 and 6 present low accuracy in finding out the root cause of the faults. In fault 21, all of the existing approaches cannot find out the root cause until the last several variables, which means diagnosing this fault needs to check a large number of variables, which will significantly increase the cost.

\begin{figure}[htbp]
  \centering
  \subfigure[0-15h]{\includegraphics[scale=0.5]{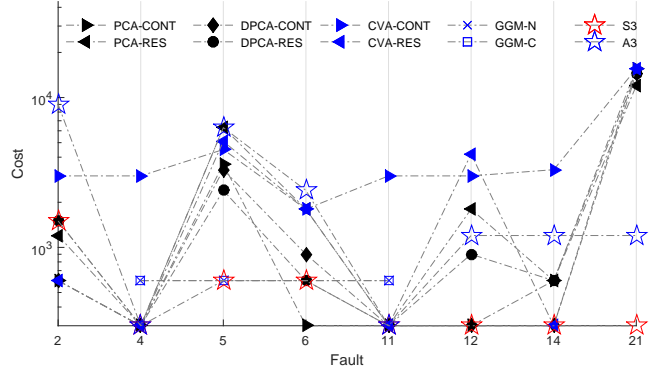}} \\
  \subfigure[15-40h]{\includegraphics[scale=0.5]{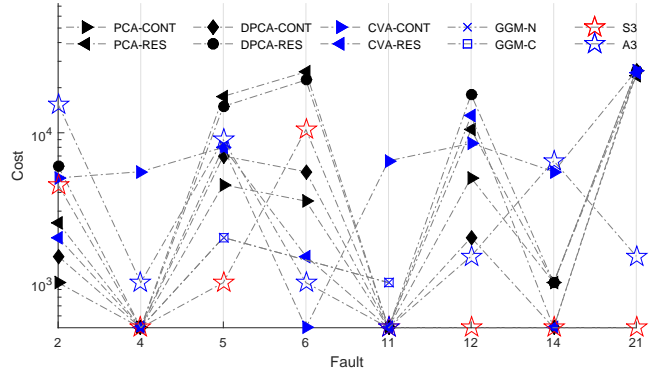}} 
  \caption{Diagnosis cost of each fault using methods listed in Table~\ref{tabMethodsList}. Each fault are considered independently, with diagnosis cost represented by different types of markers (plotted in log scale). The dash-dot lines are used to connect the same method in different faults.  }
  \label{figTEPcost}
\end{figure}

It should be noted that, \cite{russell2000data} lists the variables assumed to be most closely related to the faults, while it can be clearly observed from the raw data that the variables with significant deviations from the nominal condition are more than one in each fault. Thus it is difficult to label all the failed variables due to the lack of domain knowledge. Therefore, metrics used in Table \ref{tabNodeAccu} cannot be applied in this case. Instead, we compare the cost of identifying the root cause. We define diagnosis cost $D_c$ to evaluate the efforts to the root cause, according to the rankings in Table~\ref{tabTEPrank}. For example, if the ranking of the root cause by method $A$ is $10$, we have to check the first 10 variables ranked by method $A$ to ascertain the root cause, and we can note the diagnosis cost is $10$. By multiplying the number of measurements in each fault (treated as the number of times to implement root-cause analysis), the diagnosis cost for each fault is obtained and shown in Fig.~\ref{figTEPcost}.

Fig.~\ref{figTEPcost} shows that the proposed approach obtains better root-cause analysis results in most of the faults, compared with the other methods. GGM-N and GGM-C in \cite{qiu2012granger} only list the results of faults 4, 5, and 11 in the two time spans (6 cases in total), and $S^{3}$ and $A^{3}$ outperform 4 and 3 cases respectively. Compared with the other methods (PCA, DPCA, and CVA) in 8 faults, $S^{3}$ achieves the best performance in 6 faults whereas $A^{3}$ does not obtain the best results although it can still find out the fault variable at a reasonable cost. The diagnosis cost of the proposed approaches fluctuates less over multiple considered cases compared to other methods, meaning that our approach is robust under different scenarios. This observation is also reflected where other methods perform well in some cases, but not for some others (e.g., fault 21 in panels (a) and (b)).

\subsection{Discussions}
The proposed approaches ($S^{3}$ and $A^{3}$) in the previous sections are pattern-based root-cause methods, as opposed to node-based. With this setup, the approaches can be applied to handle both pattern and node failures. The proposed approaches present advantages in:
\begin{enumerate}
\item \emph{Ability to handle multiple nominal modes:} The STPN+RBM framework is capable of learning multiple modes and treated as nominal, which corresponds to diverse operation modes in complex CPSs. The proposed root-cause analysis approaches inherit this benefit and are validated with \texttt{dataset1}.
\item \emph{Accuracy:} The proposed approaches--$S^3$ and $A^3$--demonstrate high accuracy in root-causes analysis with synthetic data and real data. Moreover, the approach shows considerable credibility in recall, precision and F-measure in both pattern-based and node-based root-cause analysis.
\item \emph{Scalability:} The proposed approaches--$S^3$ and $A^3$--are scalable with the size of the system, and this is validated in 
TEP dataset. The scalability is important in complex CPSs where a large number of sub-systems are usually involved.
\item \emph{Robustness:} Compared with the state-of-the-art methods, the proposed approaches can more effectively isolate the fault node in both synthetic data and real data.
\item \emph{Adaptivity:} The proposed tool-chain is adaptive in different fault scenarios including pattern-based faults
    , node-based faults
    , and various fault mechanisms.
\item \emph{Efficiency:} The proposed framework is efficient in terms of diagnosis cost to ascertain the root cause for decision making and other applications.
\end{enumerate}

Note that, the root-cause analysis is only conducted when there is an anomaly detected. However, for real-life complex system monitoring, it is almost impossible to avoid false alarms completely. An experiment is carried out here where we perform root-cause analysis on 1900 nominal test cases (with 25 patterns in each case) based on the synthetic \texttt{dataset 1} (Fig. 3). We find that $S^3$ and $A^3$ algorithms detect most of the patterns to be completely nominal (93.35\% and 98.70\% respectively). Therefore, even with falsely detected anomalies, the system operator will only have to inspect a relatively smaller percentage of the patterns (6.65\% in the case of $S^3$ and 1.30\% in the case of $A^3$ for the false alarms). While we aim to reduce the false alarm rate of anomaly detection, we emphasize more on reducing the missed detection (an anomaly is not flagged when in reality it is present, i.e., false negative) rate. $S^3$ is designed with the intention of avoiding false negatives by determining the maximal inferrable subset. This increases the chance of avoiding critical safety issues and cascading system failures.

It should be mentioned that the node inference algorithm is intended to include every node until all of the detected failed patterns can be interpreted. A threshold needs to be established to avoid detecting too many nodes as faulty based on the anomaly score. Also, optimal reasoning strategy in node failure inference including single node and multiple nodes will be included in future work.

\begin{rem}
$S^{3}$ runs in a greedy manner, although the validation results show that this algorithm performs reasonably well. However, $S^3$ may erroneously say one root-cause is more important than the other in cases where multiple root-causes are equally important. This has the consequence where the user may succumb to prioritize falsely during decision-making. While $S^3$ seems to be more comprehensive and have a better performance in most cases, $A^{3}$ is comparatively more time-efficient and will potentially have superior capability in handling cases with multiple root-causes of equal importance. The average time required for $S^3$ and $A^3$ to analyze one data point in \texttt{dataset 3} using a single core CPU is 21.3 seconds and 0.4 seconds respectively. To speed up computation, it is possible to install dedicated hardware (e.g. field programmable gated arrays, or FPGAs) for parallel processing in actual implementation.

Computational complexity. Let $T$ denote the number of the data points. With a fixed STPN structure (fixed number of symbols and fixed depth for $D$-Markov and x$D$-Markov machines), the computational cost is $\mathcal{O}(f^2T)$ (based on Eq. \ref{equProbOnline1}, we use Stirling's approximation for the factorial computations, the complexity of each input is $\mathcal{O}(T)$, and there are $f^2$ patterns in the STPN model \cite{LGJS17cps}). For the $S^3$ with the trained STPN+RBM model, the overall computational complexity is $\mathcal{O}(f^2T+f^{4}n_h)$, where $n_h$ is the number of hidden units for RBM (the computational complexity of RBM inference for each input is $\mathcal{O}(f^2+f^2\times n_h)$ based on Eq. \ref{equFreEngy}, and the sequential searching takes $f^2$ times to get the subset of failed patterns in the worst case). For the $A^3$ with the trained STPN+DNN model, the overall computational complexity is $\mathcal{O}(f^2T+\mathcal{P})$, where $\mathcal{P}$ is the number of connections in the DNN model, $\mathcal{P}=f^2\times n_{h1}+\sum_{h_i=1}^{p-1} n_{h_{i}}\times n_{h_{i+1}}$, $p$ is the number of layers.

If we also consider another perspective, the combination of the two approaches will provide us the privilege to optimize both the training process and inference process. Workflow can be developed by applying $S^{3}$ for downselecting the potential failed patterns, generating a subset of training data for $A^{3}$, and then implementing inference with $A^{3}$. Further analysis is being carried out for integrating the two approaches.
\end{rem}

\section{Conclusions}
\label{secConclusion}
For root-cause analysis in distributed complex cyber-physical systems (CPSs), this work proposed an inference based metric defined on `short' time series sequences to distinguish anomalous STPN patterns. The associated analytical result formulated the root-cause analysis problem as a minimization problem through the inference based metric, and presented two new approximate algorithms--the sequential state switching ($S^3$) and artificial anomaly association ($A^3$)--for the root-cause analysis problem. Note that the rationale for qualifying the underlying system as `distributed' is that we consider that the system is composed of many sub-systems that have their own local controllers to satisfy local objectives while aiming to achieve a system-wide performance goal via physical interactions and information exchange. Therefore, it is important to capture the individual sub-system characteristics (using atomic patterns) and their relationships (using relational patterns). The proposed approaches are validated and showed high accuracy in finding failed patterns and diagnosing the anomalous node in addition to being capable of handling multiple nominal modes in operation. We also demonstrated the scalability, robustness, adaptiveness, and efficiency of the proposed approaches with comparisons to state-of-the-art methods using synthetic data sets and real data sets under a large number of different fault scenarios.

Future work will pursue: (i) optimal reasoning strategy in node failure inference including single node and multiple nodes, (ii) integration of the sequential state switching ($S^3$) and artificial anomaly association ($A^3$) to improve the performance in terms of accuracy and computational efficiency, and (iii) detection and root-cause analysis of simultaneous multiple faults in complex CPSs.

\section{Acknowledgments}
This paper is based upon research partially supported by the National Science Foundation under Grant No. CNS-1464279.

\section{SUPPLEMENTARY MATERIAL}

\subsection{Proof of Lemma 1}
\label{proof1}
\emph{
\textbf{Lemma 1.}
Let Assumption \ref{asum3} hold and $\tilde{N}_1^{ab}=t_1\tilde{N}_{11}^{ab}$ such that $(t_1-1)k>1$.
}

\begin{prof}
As $\tilde{N}_1^{ab}=\tilde{N}_{11}^{ab}+\tilde{N}_{12}^{ab}$ and $\tilde{N}_{11}^{ab}$, $\tilde{N}_{12}^{ab}$ are within similar orders, we have $t_1-1>\zeta$, where $\zeta>0$. From Assumption \ref{asum2}, we have $k\gg1$. Thus, there exists a sufficiently large $k$ to satisfy $(t_1-1)k>1$. \hfill$\square$
\end{prof}

\subsection{Proof of Proposition 1}
\label{proof2}
\emph{
\textbf{Proposition 1.}
Let Assumptions \ref{asum1}-\ref{asum3} hold. The variation of the metrics in the anomalous condition $\delta \Big(\ln( \Lambda ^{ab}) \Big)>0$, when $\eta^a\geq1$.
}

\begin{prof}
Based on the Eq. \ref{equdeltadef}, we have,
\begin{align}
\begin{split}
\nonumber
\delta \Big( &\ln( \Lambda ^{ab}) \Big)= \ln\big(\Lambda^{ab}_{nom}\big) - \ln\big(\Lambda^{ab}_{ano}\big)\\
&\quad \quad \quad \quad  = \sum_{m=1}^{|Q^{a}|}\Big(\ln((\tilde{N}^{ab}_{m})!) - \ln((\tilde{N}^{ab}_{m}-sgn(q_m)\eta^{a})!) \\
&\quad \quad \quad \quad \quad -\ln((\tilde{N}^{ab}_{m}+N^{ab}_{m}+|\Sigma^{b}|-1)!) \\
&\quad \quad \quad \quad \quad +\ln((\tilde{N}^{ab}_{m}-sgn(q_m)\eta^{a}+N^{ab}_{m}+|\Sigma^{b}|-1)!) \\
&\quad \quad \quad  \quad \quad +\ln((\tilde{N}^{ab}_{m1}+N^{ab}_{m1})!) \\
&\quad \quad \quad  \quad \quad -\ln((\tilde{N}^{ab}_{m1}-sgn(q_m)\eta^{a}+N^{ab}_{m1})!) \\
&\quad \quad \quad \quad \quad -\ln((\tilde{N}^{ab}_{m1})!)+ \ln((\tilde{N}^{ab}_{m1}-sgn(q_m)\eta^{a})!) \Big)
\label{equProbOnline2}
\end{split}
\end{align}
where $|\Sigma^{b}|=2$ in this two-state case.

For $f=\ln (n!)$, according to the derivative property of Gamma function \cite{press1987numerical}, we have,
\begin{align}
\begin{split}
\nonumber
\frac{df}{dn}=\frac{1}{n!}\frac{d(n!)}{dn}=-\gamma+\sum_{p=1}^{n}\frac{1}{p},
\end{split}
\end{align}
where $\gamma$ is the Euler-Mascheroni constant.

\begin{align}
\begin{split}
\nonumber
&\frac{d\delta \Big( \ln( \Lambda ^{ab}) \Big)}{d\eta^a}=\sum_{m=1}^2 sgn(q_m)\Bigg(\\
&\quad \quad \quad \sum_{p=1}^{t_m\tilde{N}_{m1}^{ab}-sgn(q_m)\eta^a}\frac{1}{p}
-\sum_{p=1}^{t_m(1+k)\tilde{N}_{m1}^{ab}+1-sgn(q_m)\eta^a}\frac{1}{p}\\
&\quad \quad \quad \quad \quad +\sum_{p=1}^{(1+k)\tilde{N}_{m1}^{ab}-sgn(q_m)\eta^a}\frac{1}{p}
-\sum_{p=1}^{\tilde{N}_{m1}^{ab}-sgn(q_m)\eta^a}\frac{1}{p} \Bigg)\\
\end{split}
\end{align}

For Harmonic series, $\sum_{p=1}^n\frac{1}{p}\approx \ln n+\gamma$, where $\gamma$ is the Euler-Mascheroni constant.
\begin{align}
\begin{split}
\nonumber
&\frac{d\delta \Big( \ln( \Lambda ^{ab}) \Big)}{d\eta^a}\\
&\quad \quad \quad \approx \sum_{m=1}^2 sgn(q_m)\Bigg(\ln\Big(t_m\tilde{N}_{m1}^{ab}-sgn(q_m)\eta^a\Big)\\
&\quad \quad \quad \quad-\ln\Big(t_m(1+k)\tilde{N}_{m1}^{ab}+1-sgn(q_m)\eta^a\Big)\\
&\quad \quad \quad \quad+\ln\Big((1+k)\tilde{N}_{m1}^{ab}-sgn(q_m)\eta^a\Big)\\
&\quad \quad \quad \quad-\ln\Big(\tilde{N}_{m1}^{ab}-sgn(q_m)\eta^a\Big)\Bigg)\\
&\quad \quad \quad =\ln\frac{t_1\tilde{N}_{11}^{ab}-\eta^a}{t_1(1+k)\tilde{N}_{11}^{ab}+1-\eta^a}\ \frac{(1+k)\tilde{N}_{11}^{ab}-\eta^a}{\tilde{N}_{11}^{ab}-\eta^a}\\
&\quad \quad \quad \quad +\ln\frac{t_2(1+k)\tilde{N}_{21}^{ab}+1+\eta^a}{t_2\tilde{N}_{21}^{ab}+\eta^a}\ \frac{\tilde{N}_{21}^{ab}+\eta^a}{(1+k)\tilde{N}_{21}^{ab}+\eta^a}\\
&\quad \quad \quad =\ln\frac{A_1-\tilde{N}_{11}^{ab}\eta^a(1+t_1+k)} {A_1-\tilde{N}_{11}^{ab}\eta^a(1+t_1+t_{1}k-\frac{1}{\eta^a}+\frac{1}{\tilde{N}_{11}^{ab}})}\\
&\quad \quad \quad \quad +\ln\frac{A_2+\tilde{N}_{21}^{ab}\eta^a(1+t_2+t_{2}k+\frac{1}{\eta^a})} {A_2+\tilde{N}_{21}^{ab}\eta^a(1+t_2+ k)}\\
\end{split}
\end{align}
where $A_m=t_m(1+k)\big(\tilde{N}_{m1}^{ab}\big)^2+\big(\eta^a\big)^2>0, m=1,2$.

According to Lemma \ref{lem1}, we have $(t_1-1)k>1$, For $\eta^a\geq1$,
$(t_{1}-1)k-\frac{1}{\eta^a}+\frac{1}{\tilde{N}_{11}^{ab}}>0$,
i.e.,
$k<t_{1}k-\frac{1}{\eta^a}+\frac{1}{\tilde{N}_{11}^{ab}}$. Thus,
\begin{align}
\begin{split}
\nonumber
\ln\frac{A_1-\tilde{N}_{11}^{ab}\eta^a(1+t_1+k)} {A_1-\tilde{N}_{11}^{ab}\eta^a(1+t_1+t_{1}k-\frac{1}{\eta^a}+\frac{1}{\tilde{N}_{11}^{ab}})}>0
\end{split}
\end{align}

From the definition of $t_2$ in Section \ref{subsecformulation}, $t_{2}>1$, $t_{2}k+\frac{1}{\eta^a}>k$, thus,
\begin{align}
\begin{split}
\nonumber
\ln\frac{A_2+\tilde{N}_{21}^{ab}\eta^a(1+t_2+t_{2}k+\frac{1}{\eta^a})} {A_2+\tilde{N}_{21}^{ab}\eta^a(1+t_2+k)}>0
\end{split}
\end{align}

Therefore, we have
\begin{align}
\begin{split}
\nonumber
&\frac{d\delta \Big( \ln( \Lambda ^{ab}) \Big)}{d\eta^a}>0
\end{split}
\end{align}

From Eq. \ref{equdeltadef}, if $\eta^a=0$, $\delta \Big( \ln( \Lambda ^{ab}) \Big)=0$.

Thus we have $\delta \Big( \ln( \Lambda ^{ab}) \Big)>0$, when $\eta^a\geq1$. \hfill$\square$

\end{prof}

\bibliographystyle{model1-num-names}
\bibliography{InformationFusion_v1}

\end{document}